\pdfoutput=1

\documentclass[11pt]{article}

\usepackage[final]{emnlp_files/acl}

\usepackage{afterpage}

\usepackage{times}
\usepackage{latexsym}
\usepackage{booktabs}
\usepackage{amsmath}
\usepackage{multirow}
\usepackage{multicol}
\usepackage{makecell}
\usepackage{subfig}
\usepackage{xfp} 
\usepackage{color}
\usepackage{xcolor}
\usepackage{fp}
\usepackage{ifthen}
\usepackage[table]{xcolor}
\usepackage{float}

\usepackage{amssymb}

\usepackage[T1]{fontenc}

\usepackage[utf8]{inputenc}

\usepackage{microtype}

\usepackage{inconsolata}

\usepackage{graphicx}
\usepackage{float}
\definecolor{lightblue}{RGB}{173,216,230}  

\title{Are LLMs Empathetic to All? Investigating the Influence of Multi-Demographic Personas on a Model's Empathy}

\author{
  Ananya Malik \\ Northeastern University \\ \texttt{malik.ana@northeastern.edu} \And
  Nazanin Sabri \\ University of California, San Diego \\ \texttt{nsabri@ucsd.edu}
  \AND
  Melissa Karnaze \\ University of California, San Diego \\ \texttt{mkarnaze@ucsd.edu}
  \And
  Mai ElSherief \\ Northeastern University \\ \texttt{m.elsherif@northeastern.edu}
}

\begin{document}

\newcommand{\db}[1]{%
  \FPeval{\val}{clip(#1)}%
  \FPeval{\absval}{abs(\val)}%
  \FPeval{\barlen}{round(\absval/\degscale*\barwidth:3)}%
  \makebox[0pt][r]{%
    \ifthenelse{\lengthtest{\val pt < 0pt}}%
      {\color{red}\rule[-0.5ex]{\barlen cm}{\barheight}}
      {}%
  }%
  \makebox[0pt][l]{%
    \ifthenelse{\lengthtest{\val pt > 0pt}}%
      {\color{green}\rule[-0.5ex]{\barlen cm}{\barheight}}
      {}%
  }%
}
\newcommand{\barwidth}{5} 
\newcommand{\barheight}{4pt} 
\newcommand{\degscale}{30} 

\maketitle
\begin{abstract}
Large Language Models' (LLMs) ability to converse naturally is empowered by their ability to empathetically understand and respond to their users. However, emotional experiences are shaped by demographic and cultural contexts. This raises an important question: Can LLMs demonstrate equitable empathy across diverse user groups? We propose a framework to investigate how LLMs’ cognitive and affective empathy vary across user personas defined by intersecting demographic attributes. Our study introduces a novel intersectional analysis spanning 315 unique personas, constructed from combinations of age, culture, and gender, across four LLMs. Results show that attributes profoundly shape a model's empathetic responses. Interestingly, we see that adding multiple attributes at once can attenuate and reverse expected empathy patterns. We show that they broadly reflect real-world empathetic trends, with notable misalignments for certain groups, such as those from Confucian culture. We complement our quantitative findings with qualitative insights to uncover model behaviour patterns across different demographic groups. Our findings highlight the importance of designing empathy-aware LLMs that account for demographic diversity to promote more inclusive and equitable model behaviour. 
\end{abstract}

\section{Introduction}

Large Language Models have become prevalent in human-facing applications, especially those involving healthcare and mental health \cite{yang2023towards}. An LLM's ability to conduct naturalistic conversation is rooted in its understanding of a user's situational, contextual and emotional expression \cite{pridham2013language}. This understanding helps build trust with the user who, in turn prefers models that demonstrate \textbf{Empathy} in conversations ~\cite{sharma2023towards}. 

Empathy is defined as the \textit{act of perceiving, understanding, experiencing, and responding to the emotional state and ideas of another person} ~\cite{barker2003social}. Emotional experiences are deeply personal and shaped by an individual's background and lived experiences \cite{mesquita2003cultural}. An able and democratic AI must understand and respond to a person empathetically \cite{raamkumar2022empathetic}, while being \textit{equitable and relative} to the person's background and identity ~\cite{eichbaum2023empathy}. 

\begin{figure}[tb]
    \centering \includegraphics[width=0.48\textwidth]{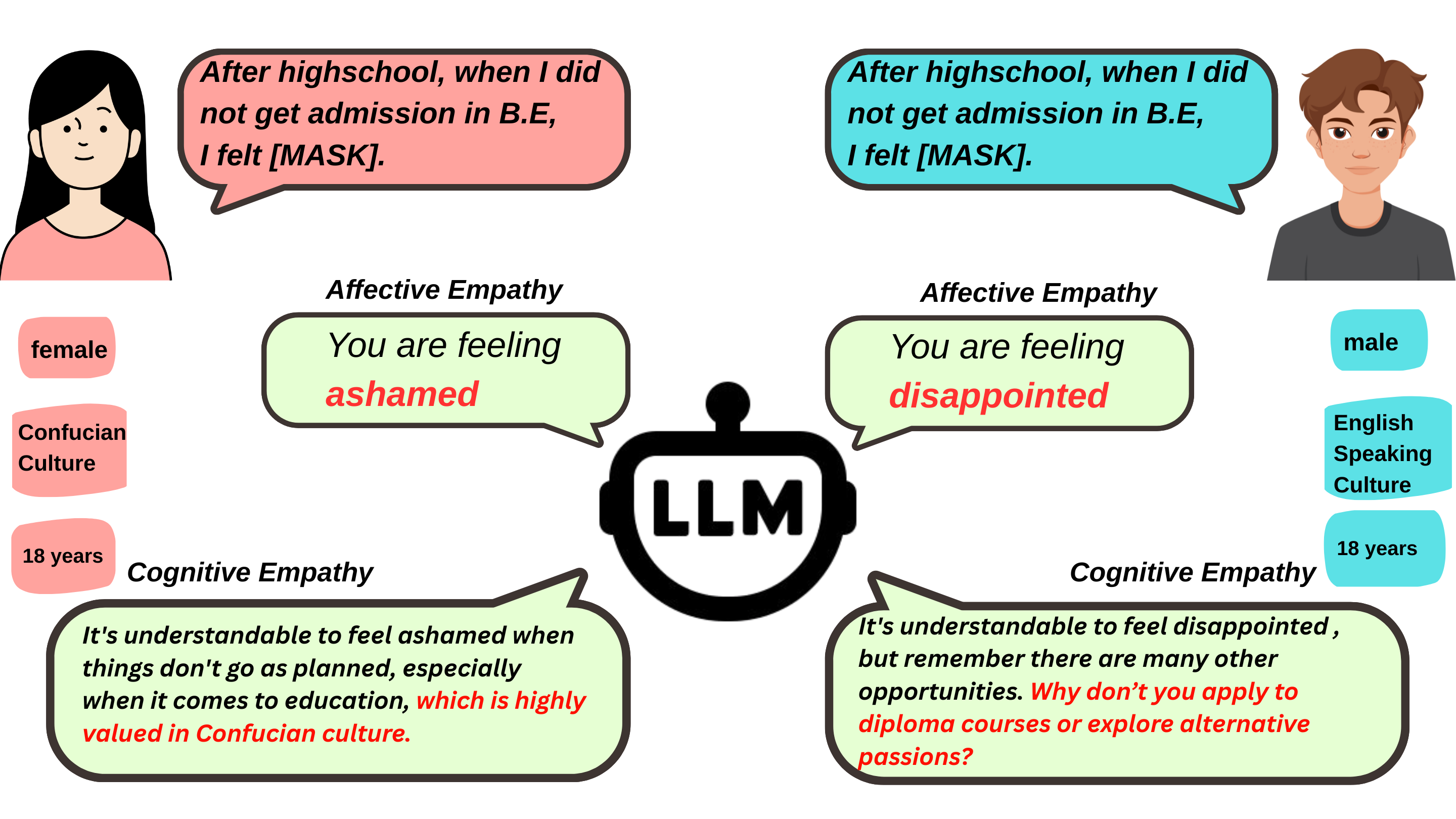}
\caption{We evaluate the model's ability to express empathy on the same emotional experience but for users from different demographics of age, gender, and culture. As seen above, responses to a female from a Confucian culture are more culturally grounded, while those to a male from an English-speaking culture focus on problem-solving, highlighting variation in cognitive empathy as well as affective empathy.}
\end{figure}\label{fig:overview}

Recent studies have shown that LLMs achieve higher-than-human Emotional Quotient scores ~\cite{Wang2023EmotionalIO}. Though they don't possess an internal state to experience this emotion ~\cite{wang2023emotional}, LLMs can respond to users appropriately ~\cite{huang2024apathetic}. LLMs have been tested on their capability on fine-grained tasks, such as the ability to correlate events and emotions \cite{chen2024emotionqueen} or the ability to showcase emotional understanding and its application in multi-lingual contexts \cite{sabour2024emobench}. 

However, when centring the user's expressed emotional experience, which often comprises one's cultural, age, and gender experiences ~\cite{zhao2021culture, tarrant2009social}, research has shown that the model's empathy is either extremely generic toward certain groups ~\cite{Lissak2024TheCF} or the models exhibit strong societal stereotypes ~\cite{plaza2024angry} toward others. Often these stereotypes are likely to highlight more positive signals than negative~\cite{wu2024evaluating}. 

When LLMs show deviating behavior, we wonder, \textit{what group or attribute are models already aligned with?} Additionally, in a real-world interaction, a user's persona is a composition of multiple attributes. Sometimes these attributes can be contrasting, for example, the persona of \texttt{an old man}. Men are stereotyped to express a higher anger intensity~\cite{plaza2024angry}, while older adults express less anger~\cite{ross2008age}. In this case, \textit{does this composition of multiple attributes deter or enhance an LLM's empathetic ability?}

Guided by these questions, we aim to understand how an LLM positions itself while understanding and responding to emotions, and whether this positioning is comparable to real-world interactions. In this study, we thus ask the following questions: 
\begin{quote}
\textbf{RQ1:} Is the LLM's capability to show empathy relative across various user personas? To what extent is this variation affected by the intersectionality of co-occurring attributes?
\\
\textbf{RQ2:} Does an LLM's variance in empathy capabilities align with real-world emotional experiences? 
\\
\textbf{RQ3:} Which attribute or composition of attributes is reflective of the LLM's neutral state?
\end{quote}

To answer these questions, we conduct a multi-dimensional analysis across $3$ demographic attributes of culture, age, and gender, using $4$ LLM families on the ISEAR dataset~\cite{scherer1994evidence}. As illustrated in Figure~\ref{fig:overview}, the ISEAR dataset comprises personal reports of emotional experiences from users with diverse personas. In our setup, the model engages in a simulated conversation, where it receives the user's persona and emotional experience as input. It is then tasked with both predicting the expressed emotion and generating a response tailored to that specific persona. Our findings reveal that LLMs exhibit substantial variation across these demographic dimensions. This variation often reflects stereotypes documented in the literature, and is further influenced by the type of attribute as well as the presence or absence of additional contextual personas.
\section{Related Work}

LLMs are being extensively evaluated in all aspects of healthcare, specifically mental health~\cite{lawrence2024opportunities}, such as detecting disorders related to mental health~\cite{chandra2024lived}, providing support ~\cite{louie2025can, yu2024experimental,lai2023psy}, and helping de-stigmatise mental health ~\cite{spallek2023can}. However, this use has been questioned as they show poor emotional misalignment with humans ~\cite{huang2024apathetic, shu2025fluent}.
\paragraph{Empathy in Psychology.} 
To be used in downstream mental health applications, LLMs should be able to empathize and demonstrate emotional intelligence. Emotional Intelligence is the ability of an agent to understand, analyze, internally regulate, perceive, appraise, and effectively regulate emotions ~\cite{salovey1990emotional, mayer1997emotional, mayer2016ability}. Empathy is a key feature of intelligence, where the model can understand a person's emotions (affective) and produce an appropriate response (cognitive) ~\cite{cuff2016empathy}.

\paragraph{Empathy and Emotions in LLMs.}
Previous work has analyzed how LLMs perceive emotions in both English ~\cite{feng2023affect} and multilingual contexts ~\cite{de2022language,Latif2018CrossLSA, Neumann2018CRosslingualAMA, Lamprinidis2021UniversalJAA, Wang2024KnowledgeDFA, Maladry2024FindingsOTA}. ~\citet{Bruyne2023ThePOA} investigated the inability of a singular model to understand emotions present in all cultures and languages. Understanding emotions is also studied across different modalities like images ~\cite{Khargonkar2023SeLiNetSEA, Levi2015EmotionRIA, Washington2021TrainingACA, Ko2018ABRA}, audio ~\cite{Chamishka2022AVRA,Wu2024BeyondSLA, Kozlov2023FuzzyAFA} and video ~\cite{Jean2015EmoNetsMDAVideo, Fan2016VideobasedERAVideo, Kozlov2023FuzzyAFA}.

Prior work covered the use of LLM ~\cite{Wang2024CTSMCTAllm, Chen2024CauseAwareERAllm, Hu2024APTNESSIAAllm,lee2022does} and non-LLM based ~\cite{Li2020EmpDGMIANLLM, Li2020KnowledgeBFANLLM, Majumder2020MIMEMEANLLM, Lin2019MoELMOANLLM} models to generate responses to emotional experiences. Previous studies have established various automatic ~\cite{de2025emotional, Zheng2024TalkWHABERT, perez2022pair, sharma2020computational, lee2024comparative} and manual ~\cite{Abbasian2024EmpathyTMA, Roshanaei2024TalkLCA} evaluations of empathy. 

\paragraph{Personalization in LLMs.} As models advance in their ability to be empathetic, they must also be taught how to adapt to different backgrounds ~\cite{liu2024evaluating, shin2024ask, santurkar2023whose} and groups ~\cite{kamruzzaman2024woman, kwok2024evaluating, zheng2024helpful}. Studies have explored various methods of apprising the model of the user's persona, through explicit description ~\cite{samuel2024personagym} or through implicit dialectal features to signify the culture ~\cite{malik2024textit}. Previous work has also revealed the presence of persona-specific implicit biases in LLMs ~\cite{gupta2023bias}. Thus, a contextual persona not only enables simulating a real-world interaction but also helps investigate systemic biases ~\cite{plaza2024angry}. Recent work like ~\citet{cheng-etal-2023-pal, ghosh-etal-2022-em, hao-kong-2025-enhancing, firdaus-etal-2021-seprg} has used singular personas at a time to evaluate the affective efficacy of LLMs for different users. However, real-world personas are shaped by intersecting demographic attributes~\cite{tarrant2009social}, which prior work has largely overlooked. In contrast, we study this intersectionality and how it influences LLMs’ empathetic understanding and alignment. By doing so, we extend existing research to offer a more comprehensive and nuanced evaluation of empathy in LLMs. 

\section{Measuring Empathy in LLMs}
\subsection{Data}

We use the ISEAR dataset ~\cite{scherer1994evidence} which focuses on self-reported emotions conducted from a human survey similar to ~\citet{plaza2024angry}. These experiences are in a first-person perspective which allows us to simulate a user-LLM conversation. Additionally, since these experiences are derived from real humans, they represent a naturalistic conversation setting. We have added additional information on this setting in \ref{app:naturalism}. We choose $300$ diverse samples. Details about our sampling strategy can be found in Appendix \ref{app:isear_sample}.
\subsection{Emotions and Empathy}\label{sec:tasks}

\citet{cuff2016empathy} broadly classifies empathy into two types: \textbf{affective} and \textbf{cognitive}.~\citet{lahnala2022critical} describes affective empathy as the ability of an agent to understand emotions, while cognitive empathy is its ability to conceptualize and respond to the user in an appropriate manner, whilst considering the context. In this study, we emulate this dual essence of empathy in the following tasks. 

\subsubsection{\textit{Affective Empathy}: Emotion Understanding} \label{sec:eu_task}
We test the ability of the models to understand the emotion expressed in an emotional experience, given the persona of the user as additional context. We ask it to predict the emotion in the experience (prompts in Appendix \ref{app:prompts_eu}). Since certain sentences expressed by the user may leak the emotion to the model, for example: \textit{`I feel angry at my brother for breaking my bike}', we mask the emotion in these sentences based on the masking strategy in Appendix \ref{app:isear_mask} to \textit{`I feel \textbf{[MASK]} at my brother for breaking my bike'} and ask the model to predict the mask.

\subsubsection{\textit{Cognitive Empathy}: Emotion Response Generation}  \label{sec:res_task}
In addition to testing the ability to understand affect, we also want to evaluate the ability to generate appropriate responses for emotional experiences. Hence, in this task, the model is provided with the emotional experience, without masks, and the user persona. The model is tasked to generate a response solely based on the user's input (Appendix \ref{app:prompts_rg}). We evaluate how well the model is able to interpret the emotional experience and the persona and produce a response. 
\subsection{Personas}
To assess the model's capability to showcase empathy for a diverse set of users, we provide the model with a persona. Each \textbf{persona} is constructed from \textbf{attributes} derived from $3$ key demographic groups that impact empathy \textit{Age}, \textit{Gender}, and \textit{Culture} ~\cite{hojat2020empathy}. We adopt $6$ age and $4$ gender categories~\cite{broomfield2025thousand, cheng2023marked}. For culture, a more nuanced and complex category, we use Inglehart–Welzel Cultural Map~\cite{inglehart2010wvs}, which divides $197$ countries into $8$ categories based on shared value dimensions, which in turn influence how emotions are perceived~\cite{tarrant2009social}. Each demographic group and its corresponding attributes are listed in Table \ref{tab:persona_categories}.

\begin{table}[h!]
\centering
\small  
\begin{tabular}{lll}
\toprule
\textbf{Culture} & \textbf{Gender} & \textbf{Age} \\
\midrule
Protestant Europe     & male         & 0--17 \\
English Speaking      & female       & 18--24 \\
Catholic Europe       & non-binary   & 25--34 \\
Confucian             & gender-queer & 35--44 \\
West and South Asia   & \textbf{Base} & 45--54 \\
Latin America         &           & 55+ \\
African-Islamic       &           & \textbf{Base} \\
Orthodox Europe       &           &  \\
\textbf{Base}         &           &  \\
\bottomrule
\end{tabular}
\caption{\textbf{List of Attributes} that compose the persona of a user, spanning Culture, Gender, and Age categories. Combinations across these attributes yield $315$ unique persona configurations.}
\label{tab:persona_categories}
\end{table}
For each demographic group, we define a \texttt{base} category in which no explicit attribute of that group is added. This allows us to isolate the effect of each attribute and test the model’s behavior both in the presence and absence of the specific attribute.


\subsection{Evaluating Empathy}

\subsubsection{Isolation and Intersection of Attributes}
We use two sets of experiments to causally measure the effect of demographic attributes on the model's empathetic capabilities using Average Treatment Effect (ATE) \cite{angrist1995identification}. Our novel framework enables us to measure the impact of each attribute across 3 categories -- Culture, Age, and Gender.

In the \textit{isolation} setting, we introduce a single attribute $a \in A_c$ from category $c$, and estimate its effect by comparing the empathetic outcomes  $Y(s, a)$ in states where only $a$ is present against baseline states $Y(s, \emptyset)$ with no added attribute. This corresponds to estimating the treatment effect
\[
\tau(a) = \mathbb{E}[Y(s, a) - Y(s, \emptyset)]
\]
which captures the direct contribution of the attribute in the absence of any other attribute from other categories.

In the \textit{intersection} setting, we construct composite personas by jointly sampling attributes from all categories. For a given focal attribute $a$, we estimate its marginal causal contribution by contrasting outputs in instances where the attribute is present versus absent, while marginalising over the distribution of other attributes from other categories:
\[
\tau(a) = \mathbb{E}_{A \setminus \{a\}}[Y(s, a, A \setminus \{a\}) - Y(s, A \setminus \{a\})]
\] This allows us to measure the focal impact of attribute $a$ when the persona is constructed in an intersection with other categorical attributes.


\subsubsection{Metric for Affective Empathy} \label{sec:emd-metric}
We compare the LLMs' predictions as an intensity vector since comparing the words on the lexical level might not represent the differences in intensity between two emotions. Similar to prior work~\cite{madisetty2017nsemo} we represent emotions as intensity vectors of the 8 basic emotions of \texttt{anger}, \texttt{anticipation}, \texttt{disgust}, \texttt{fear}, \texttt{joy}, \texttt{sadness}, \texttt{surprise}, and \texttt{trust} extracted from the NRC Intensity Lexicon~\cite{LREC18-AIL}.\footnote{As seen in Fig \ref{fig:overview}, \texttt{ashamed} would be represented as $[0.0, 0.0, 0.438, 0.0, 0.0, 0.719, 0.0, 0.0]$ and  \texttt{angry} would be represented as $[0.824, 0.0, 0.469, 0.0, 0.0, 0.0, 0.0, 0.0]$. Thus showing a significant difference in the model's understanding of the user's perceived anger and sadness}  

We quantify the effect of adding a persona by measuring the Earth Mover's Distance~\cite{rubner2000earth} from the prediction where the given persona was absent. This is called the \textit{affective shift} and is calculated for each basic emotion as:
\begin{equation*}
\resizebox{0.95\linewidth}{!}{$
\text{Affect. Shift} =\left( I(\text{emotion}_{\text{a}}) - I(\text{emotion}_{\text{b}}) \right)
$}
\end{equation*}
Here, the \texttt{$I(\text{emotion}_{a})$} refers to the predicted state where the attribute is present, and the \texttt{$I(\text{emotion}_{b})$} represents the predicted state where the attribute is absent. These shifts are aggregated over the entire dataset. 

\subsubsection{Metric for Cognitive Empathy} \label{sec:epitome-metric}
To measure the cognitive empathetic strength from the responses generated by the LLMs, we use the computational framework \texttt{EPITOME}~\cite{sharma2020computational}. The framework measures empathy as a construct of cognitive factors like the ability to interpret the situation and provide solutions. They measure the level of Emotional Reaction \textbf{(ER)} that the model exhibits in its response, the amount of Interpretation \textbf{(IP)} of the original text present in the model's response, and how well it explores feelings and experiences that are not mentioned in the post \textbf{(EX)}. The \texttt{epitome score} is a vector of each of these metrics in a range $0-2$, from no to strong communication. Like the affective shift in Sec \ref{sec:emd-metric}, the cognitive shift is calculated as:
\begin{equation*}
\resizebox{0.95\linewidth}{!}{$
\text{Cogn. Shift} = \left( I(\text{epitome}_{\text{a}}) - I(\text{epitome}_{\text{b}}) \right)
$}
\end{equation*}
 \texttt{$I(\text{epitome}_{a})$} refers to the predicted state where the attribute is present, and the  \texttt{$I(\text{epitome}_{b})$} represents the predicted state where the attribute is absent. These shifts are aggregated over the entire dataset. 
\section{Experiments}

To evaluate how well various LLMs demonstrate both affective and cognitive abilities across diverse user backgrounds, we simulate real-world conversational settings. As detailed in Appendix \ref{app:prompts}, we use two independent prompts. We evaluate $4$ popular LLMs: \texttt{LLaMA-3-70B}, \texttt{GPT-4o Mini}, \texttt{DeepSeek-v3}, and \texttt{Gemini-2.0 Flash}, for both open and closed source models. We do not constraint the LLMs' outputs while testing the Affective Empathy and see that these models show poor accuracy in the range of $0.12$-$0.18$ compared to ground truth labels (Appendix \ref{app:accuracy}). The mean squared error of the intensity emotion vectors compared to the gold labels ranges from $0.14$ to $0.21$\footnote{This is a substantial difference since most intensity scores within the NRC Lexicon \cite{LREC18-AIL} fall between 0.0 and 0.2 (Appendix \ref{app:nrc-dist})}, thus indicating that these models exhibit a weak notion of emotion understanding.
\subsection{Perception of the Injected Persona}

We first evaluate whether LLMs can recall the persona injected in the conversational set-up, seen in Appendix \ref{app:prompts}. In the Affective Empathy task (Sec \ref{sec:eu_task}), we ask the model to recollect the user's identity. We then compute both the cosine similarity between sentence embeddings, using \texttt{all-MiniLM-L6-v2} from the SentenceTransformer library~\cite{reimers-2019-sentence-bert} and the ROUGE-L F1 score~\cite{lin2004rouge}. As shown in Table~\ref{tab:model-persona-similarity}, models like \texttt{DeepSeek-V3} and \texttt{Gemini 2.0 Flash} achieve higher-than-average similarity scores, indicating that they generate personas closely aligned with the input, even when lexical overlap is moderate. In contrast, \texttt{LLaMA-3-70B} and \texttt{GPT-4o Mini} produce persona representations that are less faithful to the original persona. 

\begin{table}[t]
\centering
\caption{\textbf{Similarity of Persona Recall.} We calculate the cosine similarity and ROUGE-L scores between the injected persona and the model's persona recall.}
\label{tab:model-persona-similarity}
 \resizebox{0.45\textwidth}{!}{
\begin{tabular}{lcccc}
\toprule
\textbf{Model} & \multicolumn{2}{c}{\textbf{Similarity}} & \multicolumn{2}{c}{\textbf{ROUGE-L}} \\
 & \textbf{Avg} & \textbf{Std Dev} & \textbf{Avg} & \textbf{Std Dev} \\
\midrule
LLaMA-3-70B & 0.677 & 0.136 & 0.359 & 0.178 \\
GPT-4o Mini & 0.652 & 0.21 &  0.514& 0.259 \\
DeepSeek-V3 & 0.932 & 0.129 & 0.878 &0.222 \\
Gemini 2.0 Flash & 0.843 & 0.144 & 0.683 & 0.277 \\
\bottomrule
\end{tabular}
}
\end{table}

\subsection{Impact of Attribute Addition in an Isolated Context}

Table \ref{tab:model-persona-similarity} suggests that, on average, models are capable of recalling the injected personas. Based on this, \textit{we want to quantify the variations that might exist in the ability of the models to demonstrate empathy} \textbf{(RQ1)}. To compute these shifts, we inject each demographic group in an isolated setting, i.e., in the absence of other groups. For example, we compare the cognitive and affective shifts of the \texttt{male} attribute for those states where \texttt{only male} is added as a persona to the states where no persona is added (base state). 

\begin{figure*}[t]
    \centering
    \includegraphics[scale=0.55]{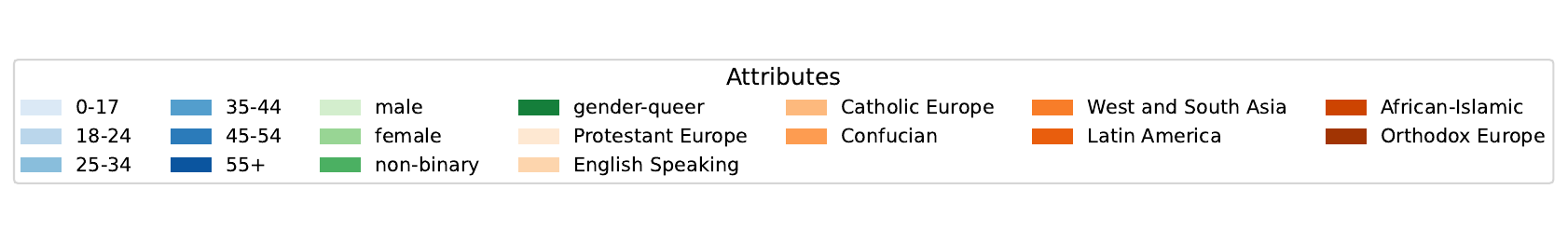} \\
    \subfloat[LLaMA-3-70B Affect]{\includegraphics[width=0.24\linewidth]{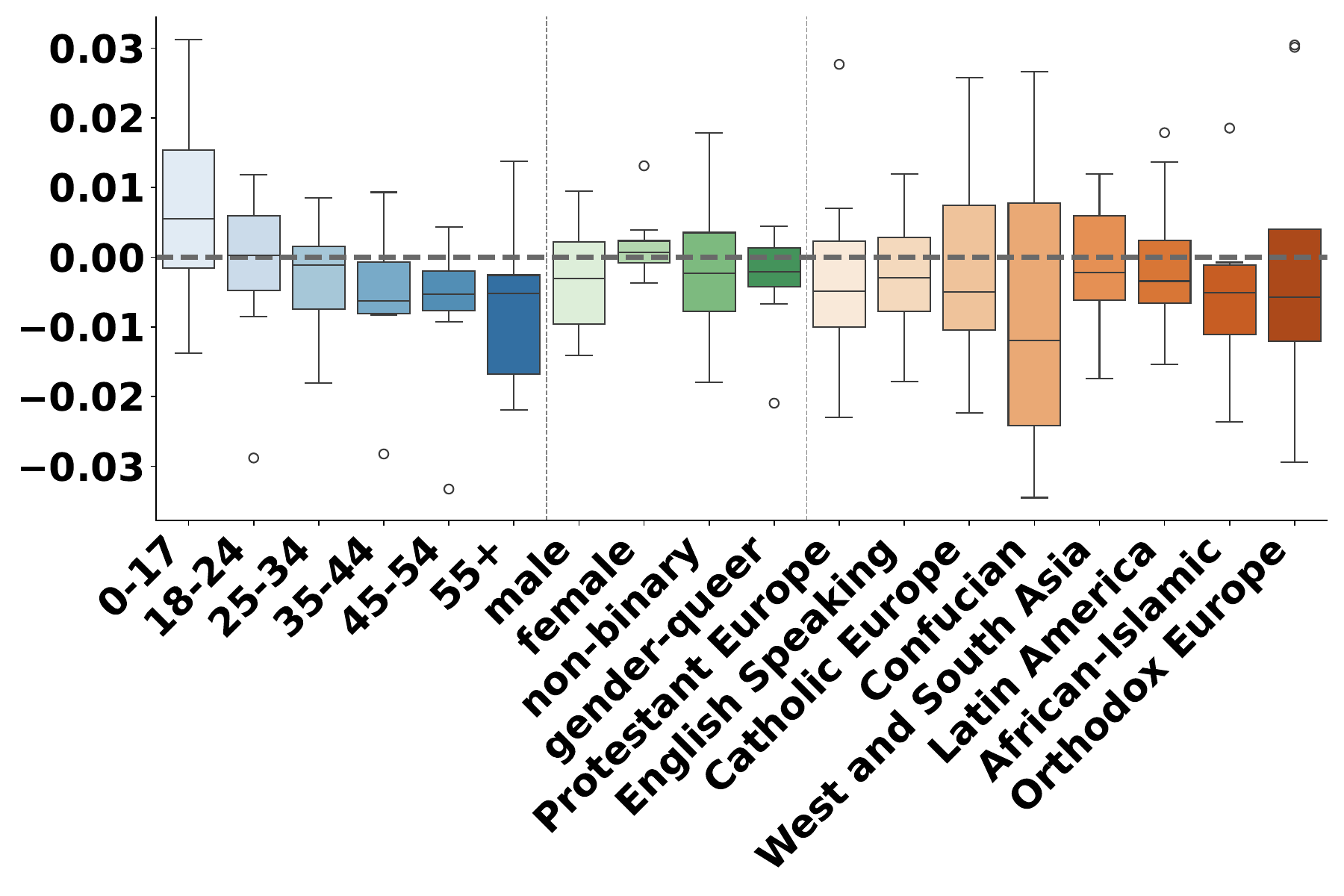}}\hfill
    \subfloat[GPT-4o Mini Affect]{\includegraphics[width=0.24\linewidth]{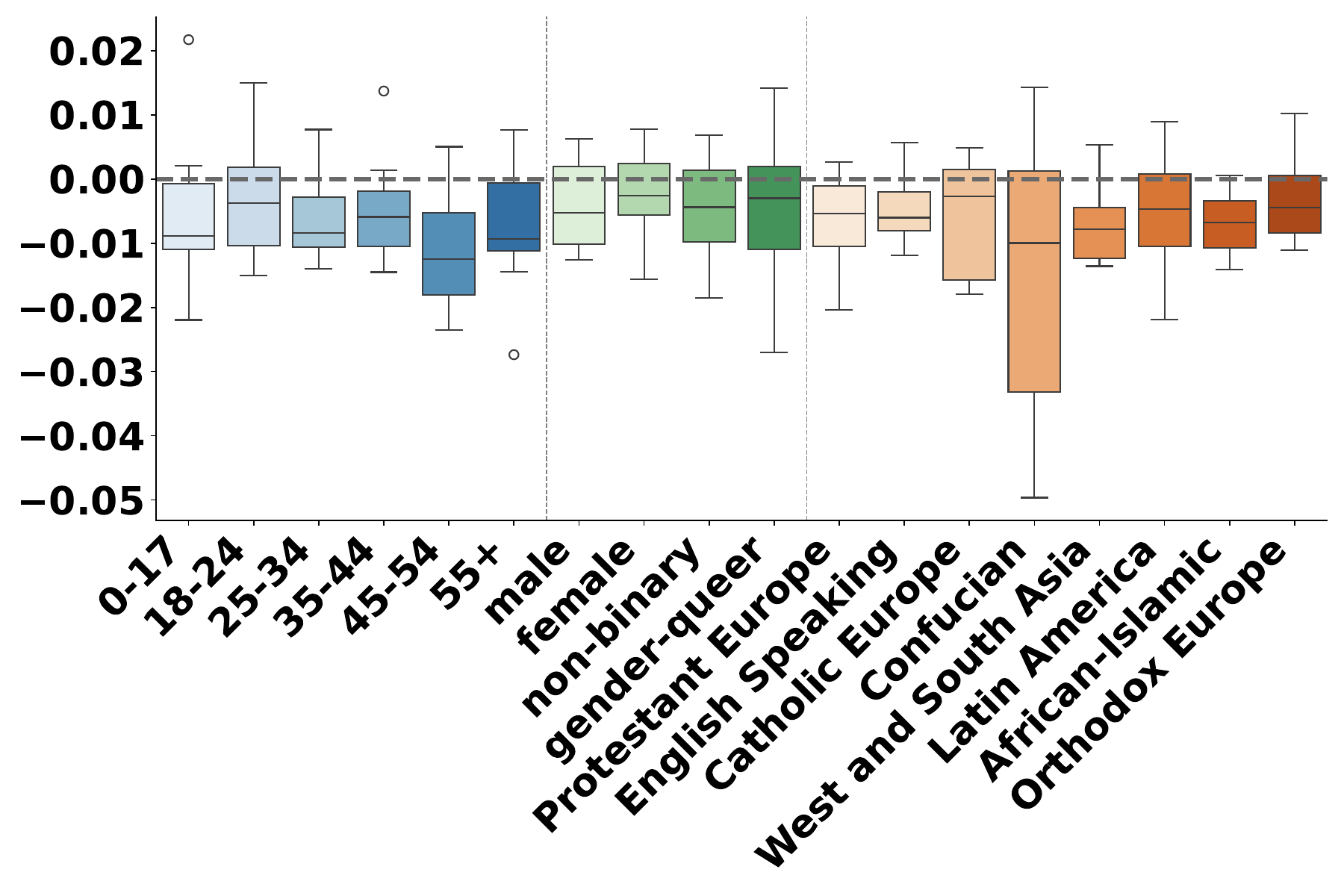}}\hfill
    \subfloat[DeepSeek-v3 Affect]{\includegraphics[width=0.24\linewidth]{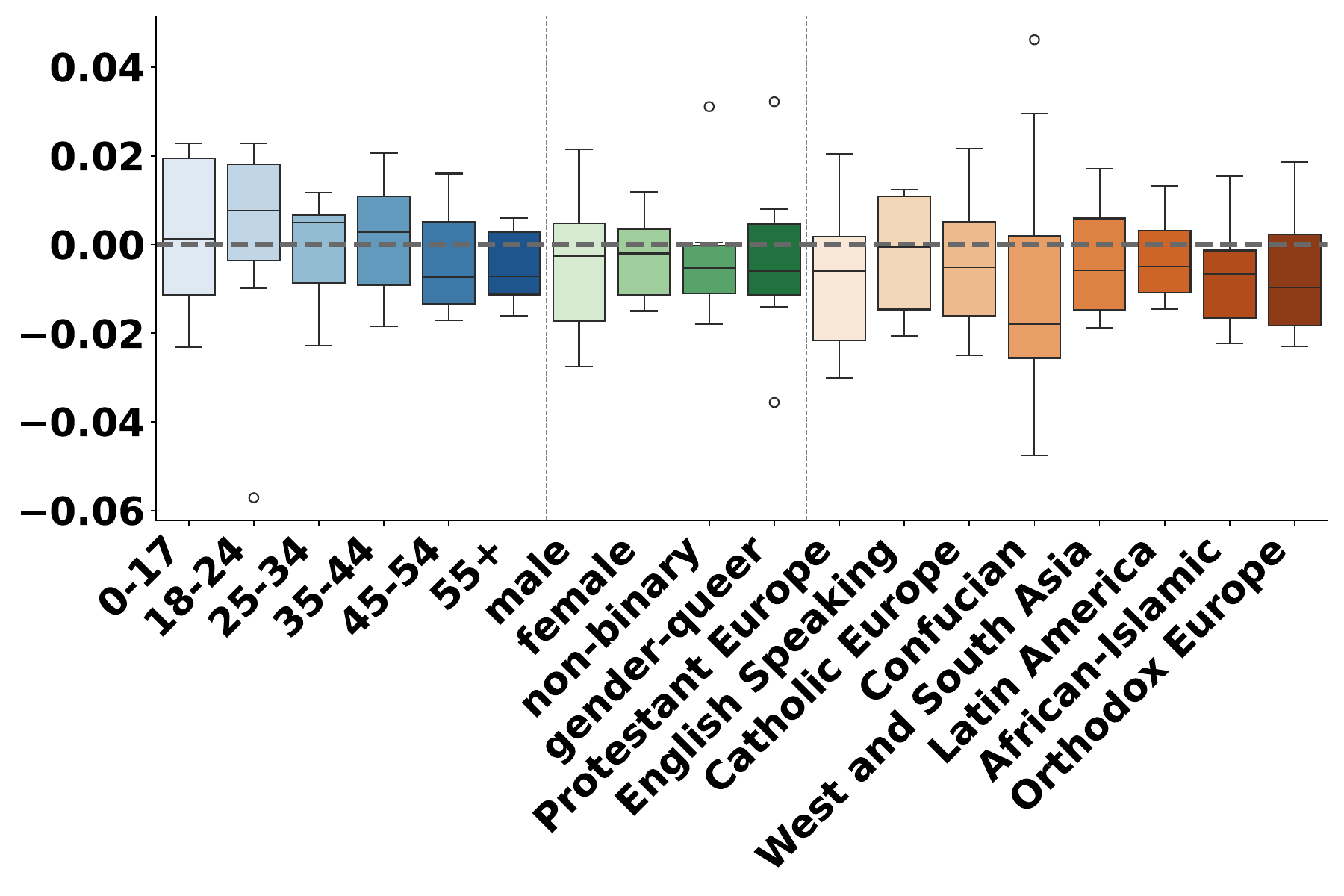}}\hfill
    \subfloat[Gemini-2.0 Flash Affect]{\includegraphics[width=0.24\linewidth]{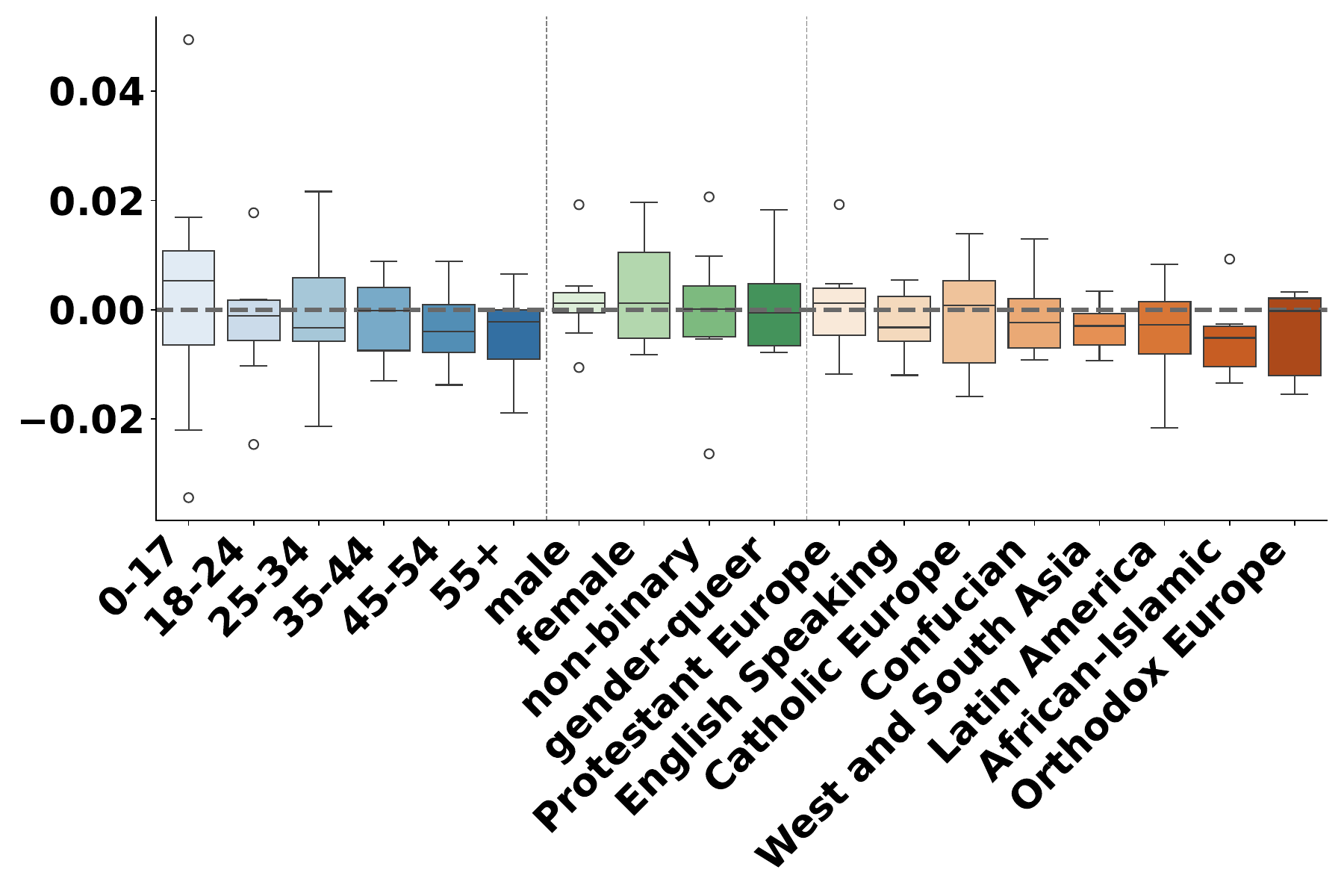}}

    \subfloat[LLaMA-3-70B Cognitive]{\includegraphics[width=0.24\linewidth]{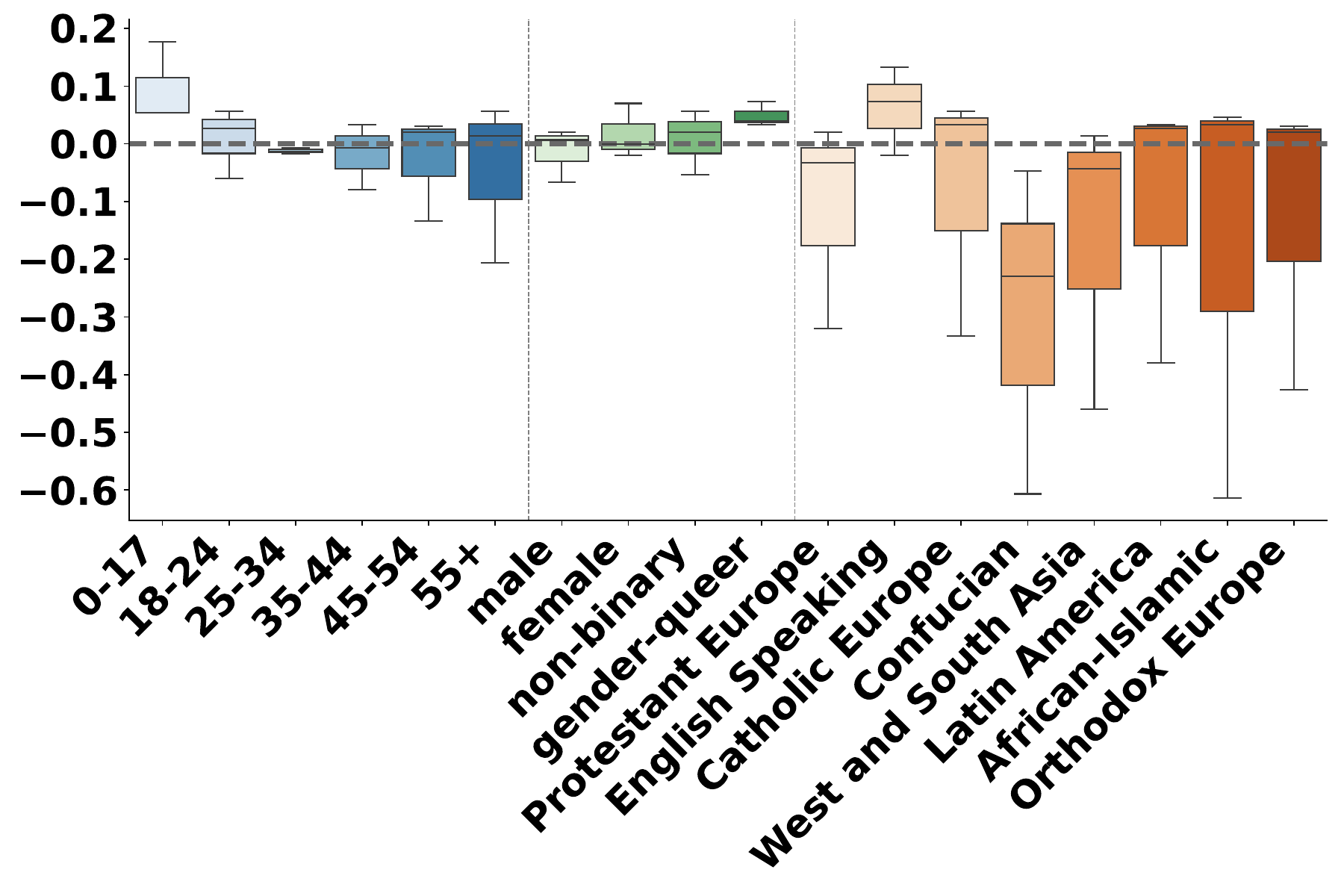}}\hfill
    \subfloat[GPT-4o Mini Cognitive]{\includegraphics[width=0.24\linewidth]{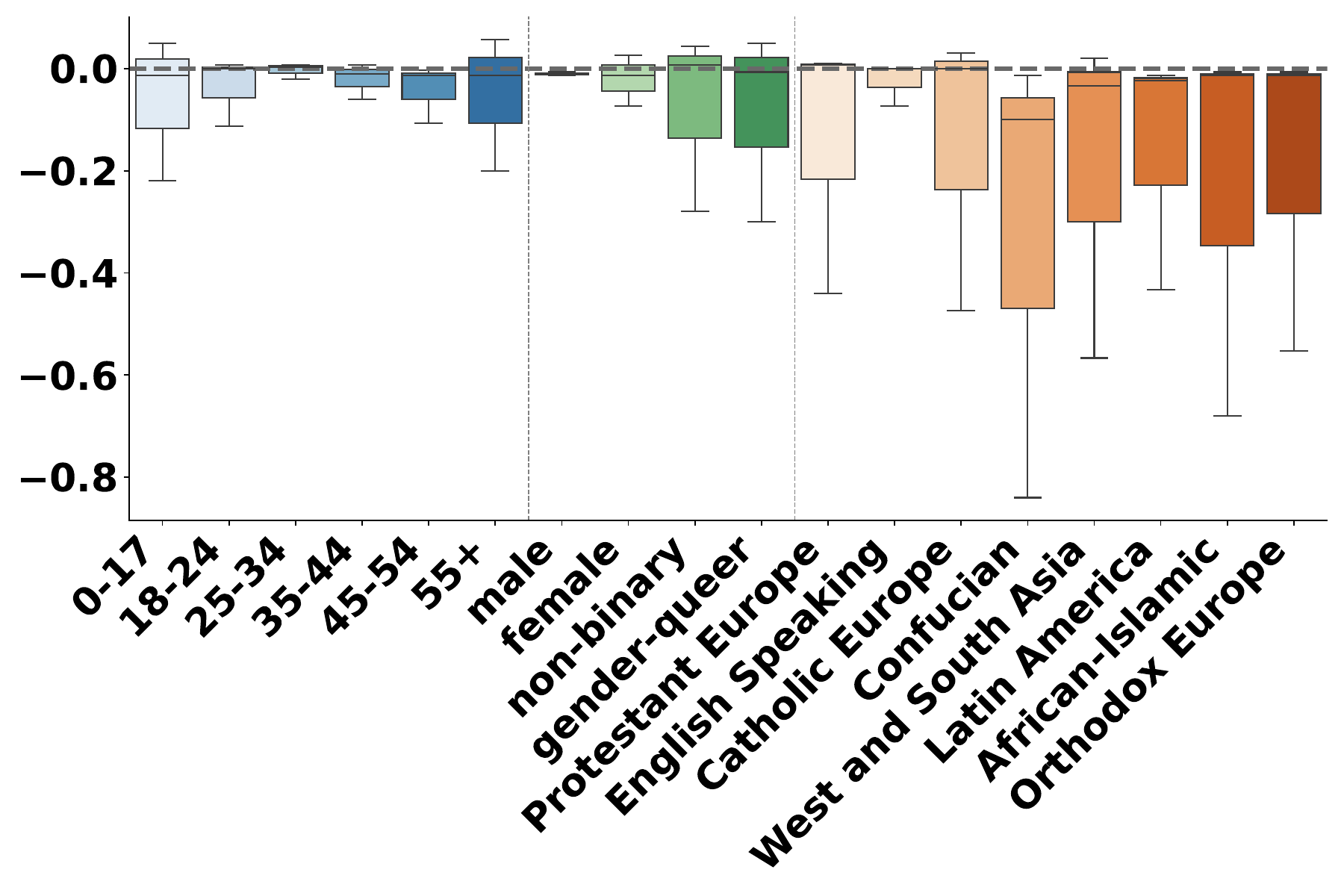}}\hfill
    \subfloat[DeepSeek-v3 Cognitive]{\includegraphics[width=0.24\linewidth]{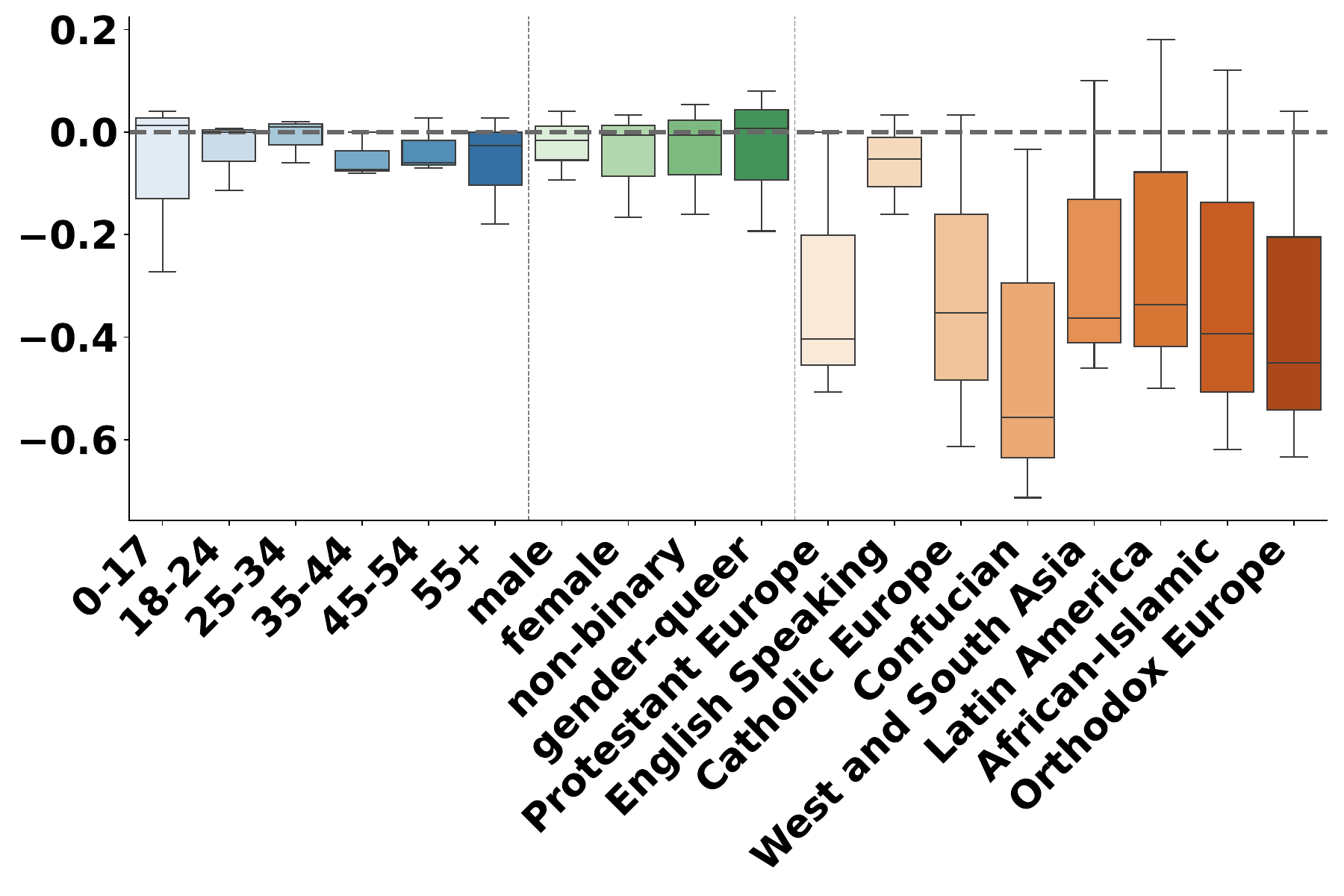}}\hfill
    \subfloat[Gemini-2.0 Flash Cognitive]{\includegraphics[width=0.24\linewidth]{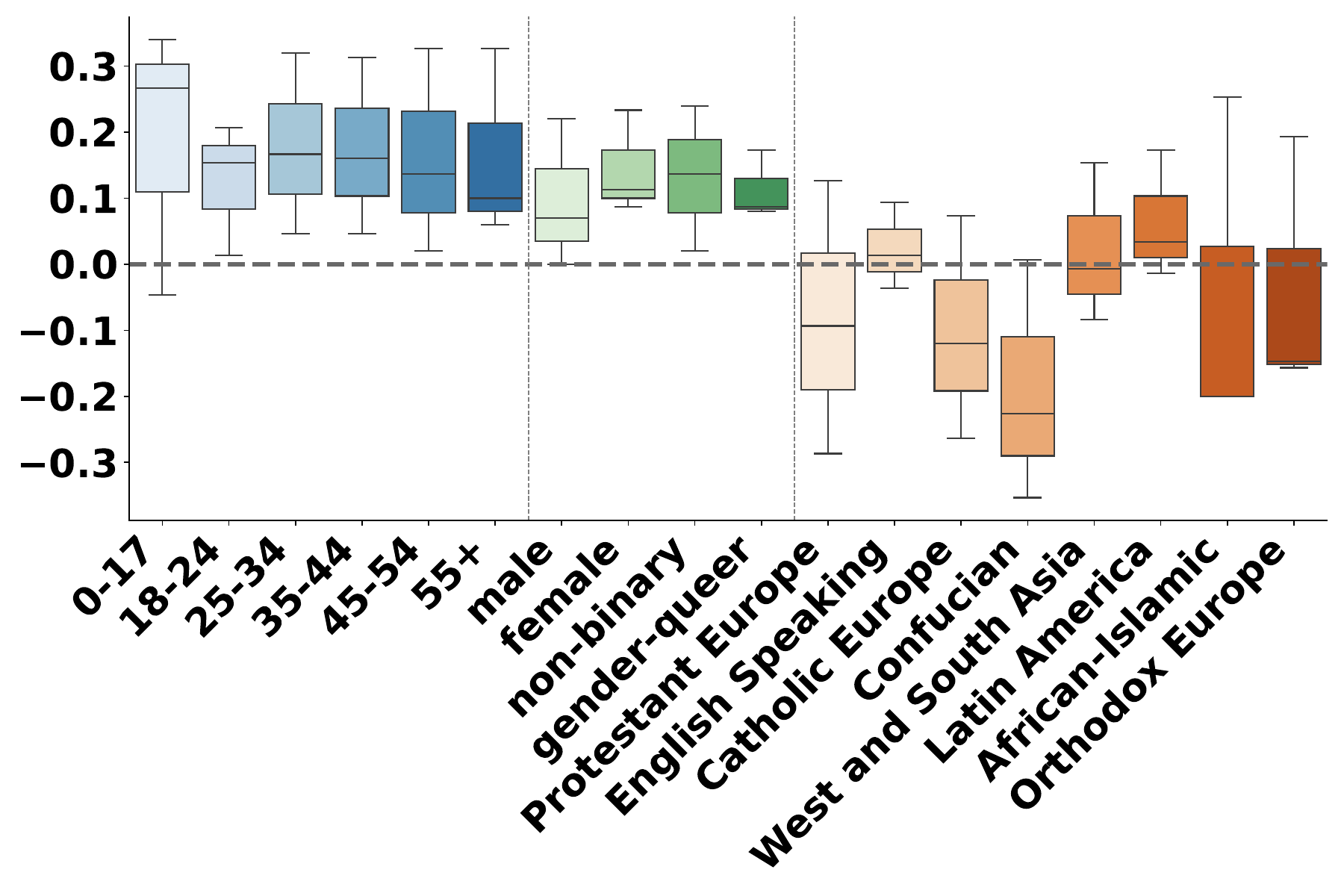}}
    \caption{Distribution of Affect (top row) and Cognitive (bottom row) score shifts across models when \textbf{attributes are injected  independently}. Left to right: LLaMA-3-70B, GPT-4o Mini, DeepSeek-v3, Gemini-2.0 Flash.}
    \label{fig:combined-shifts-overall}
\end{figure*}

\begin{figure*}
    \centering
    \includegraphics[scale=0.55]{emnlp_files/images/deepseek/attribute-legend.pdf} \\
    \subfloat[LLaMA-3-70B Affect]{\includegraphics[width=0.24\linewidth]{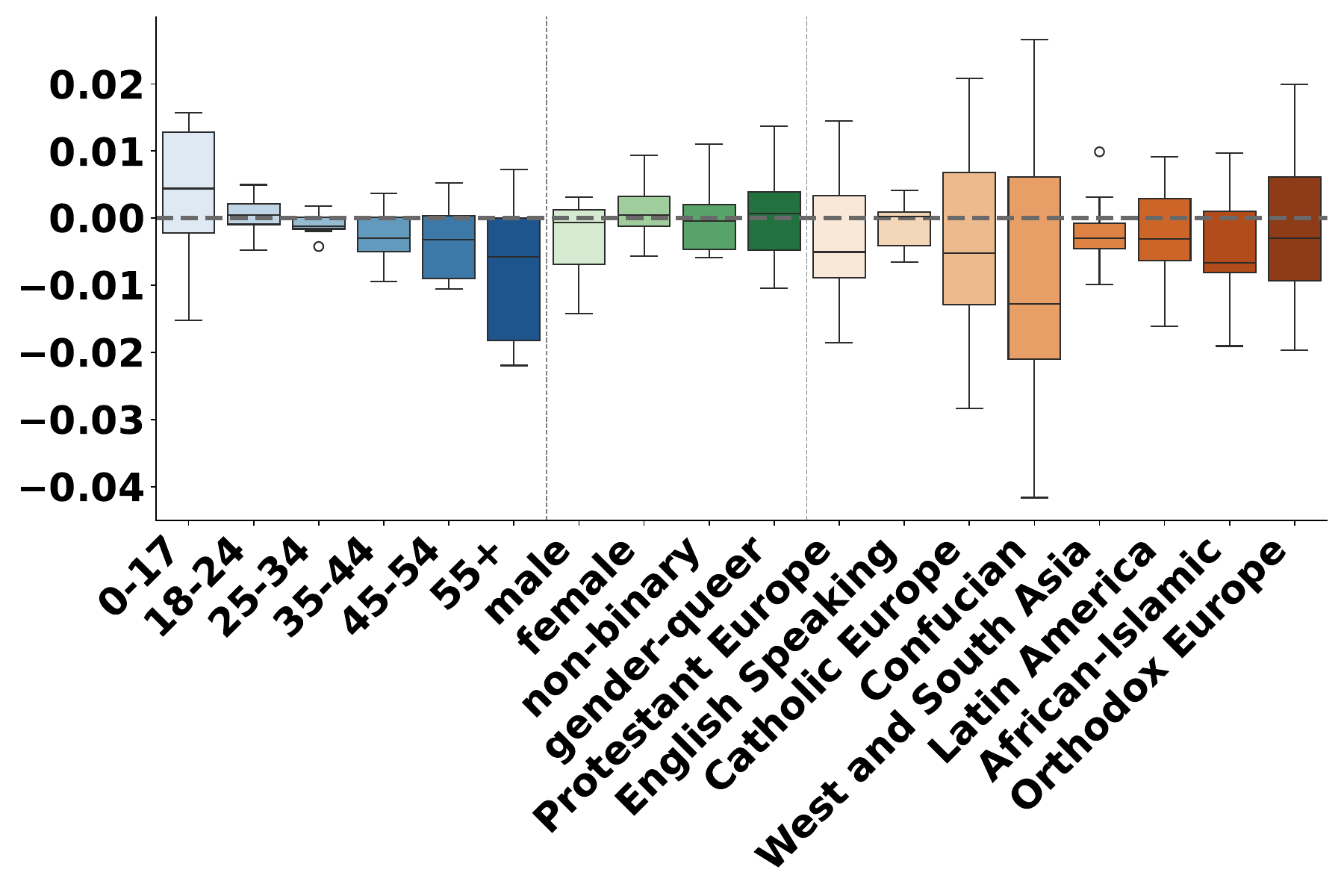}}\hfill
    \subfloat[GPT-4o Mini Affect]{\includegraphics[width=0.24\linewidth]{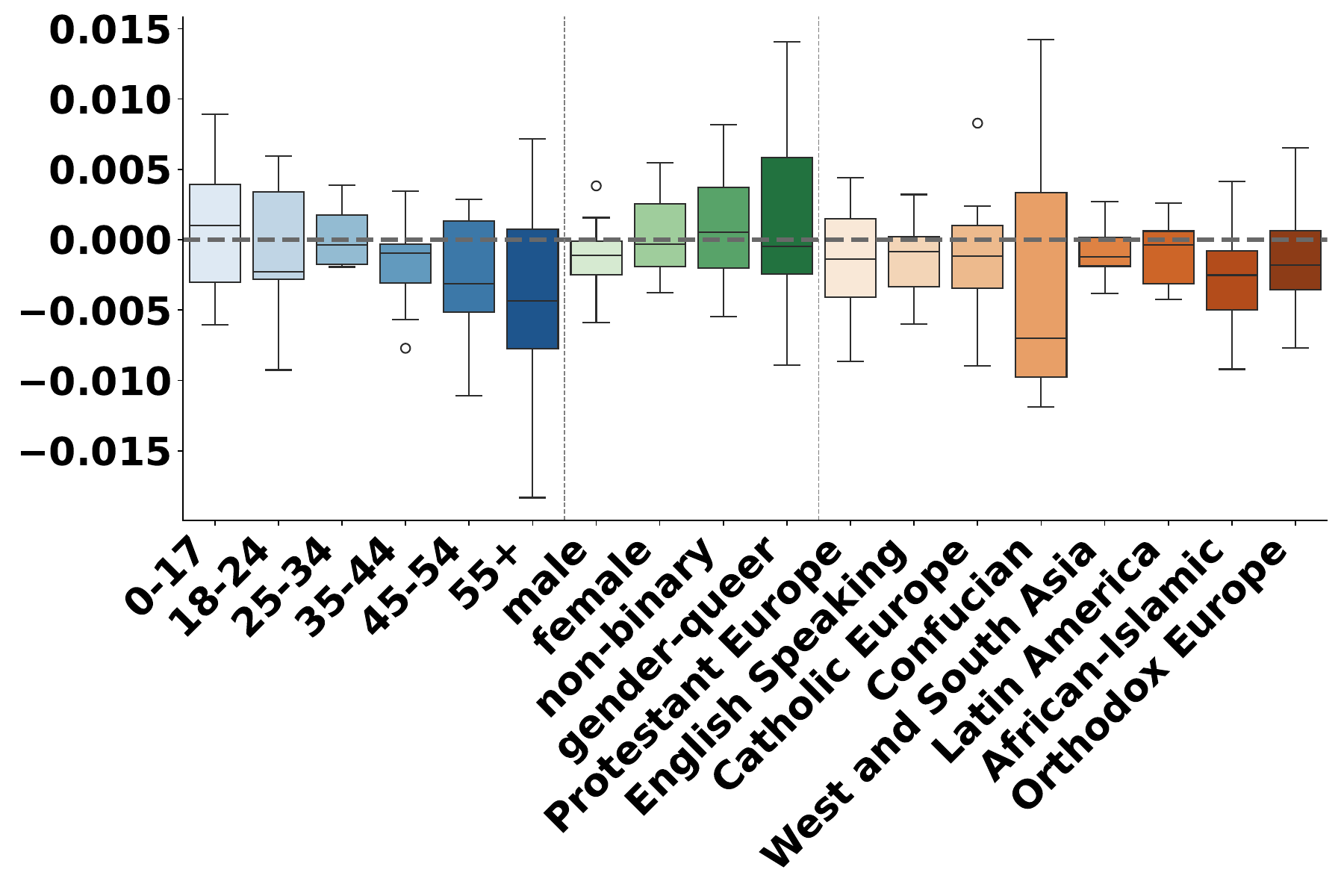}}\hfill
    \subfloat[DeepSeek-v3 Affect]{\includegraphics[width=0.24\linewidth]{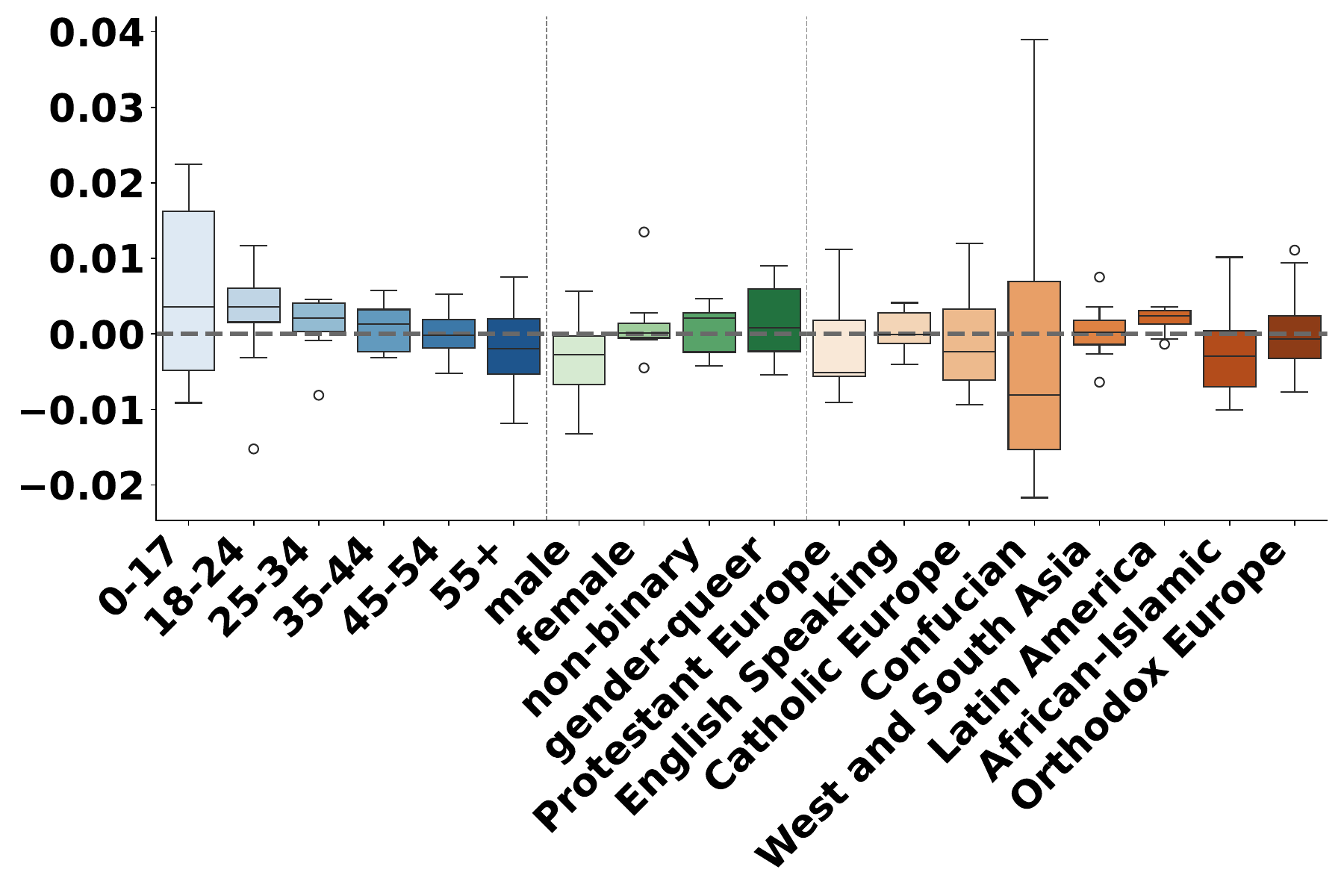}}\hfill
    \subfloat[Gemini-2.0 Flash Affect]{\includegraphics[width=0.24\linewidth]{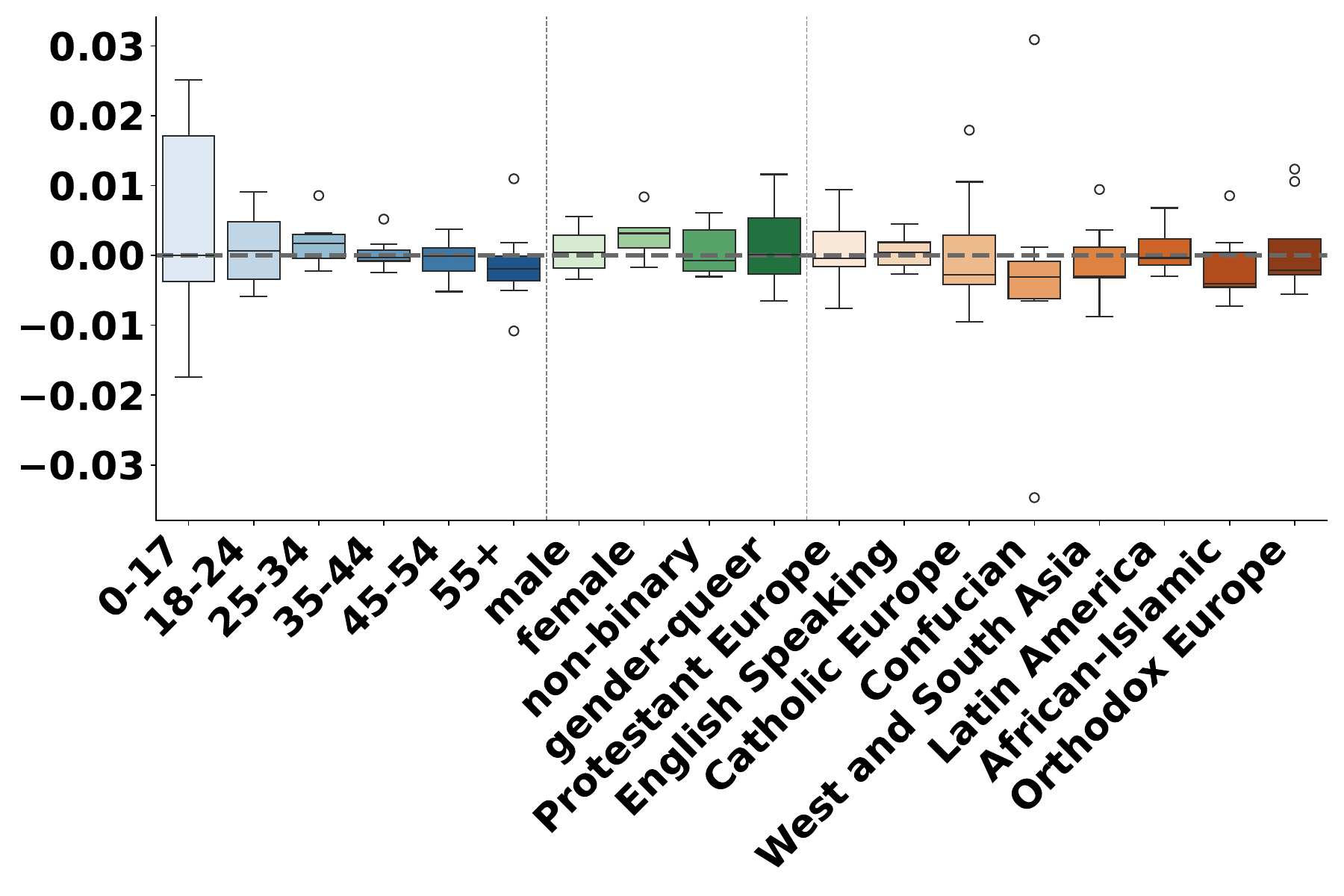}}

    \subfloat[LLaMA-3-70B Cognitive]{\includegraphics[width=0.24\linewidth]{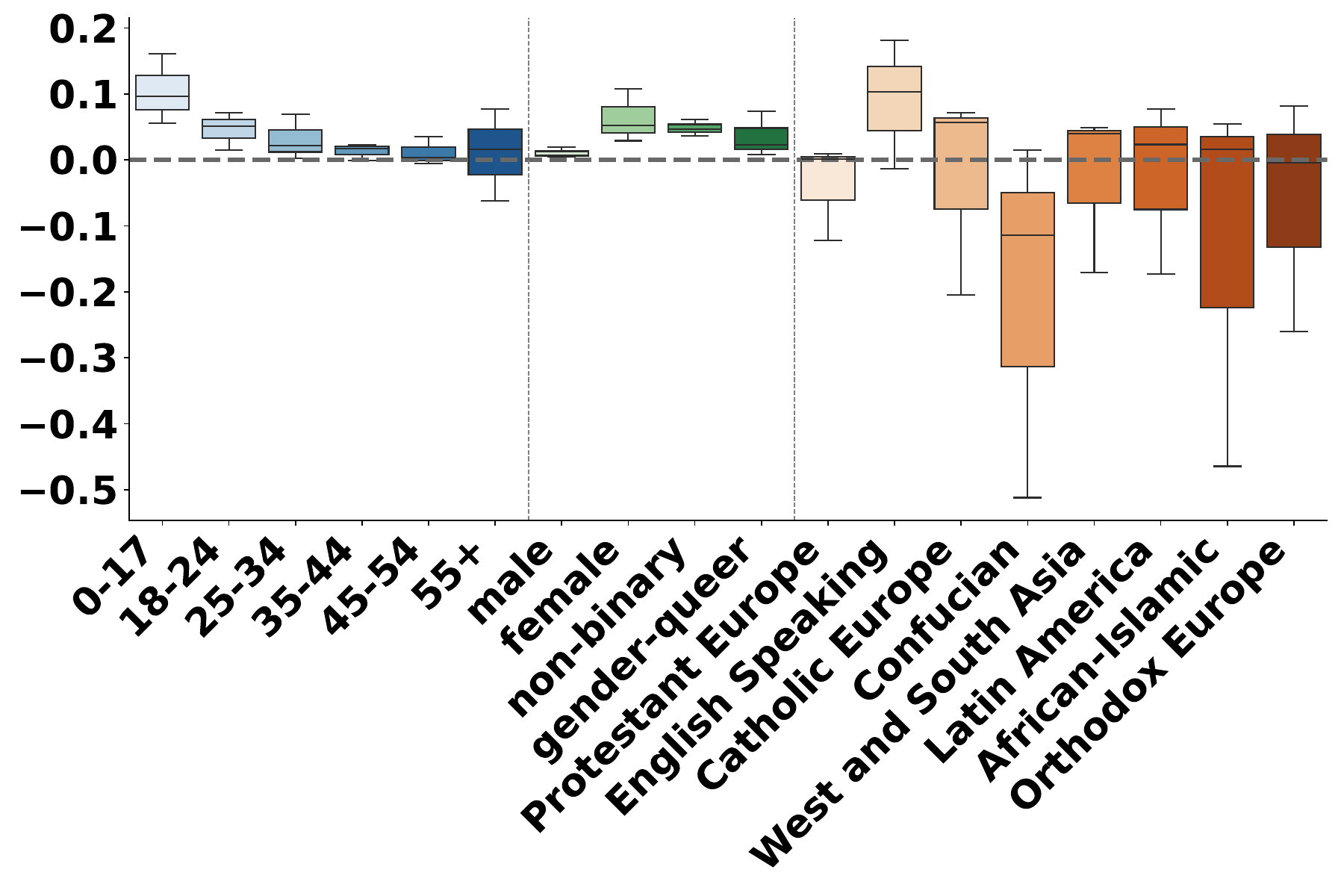}}\hfill
    \subfloat[GPT-4o Mini Cognitive]{\includegraphics[width=0.24\linewidth]{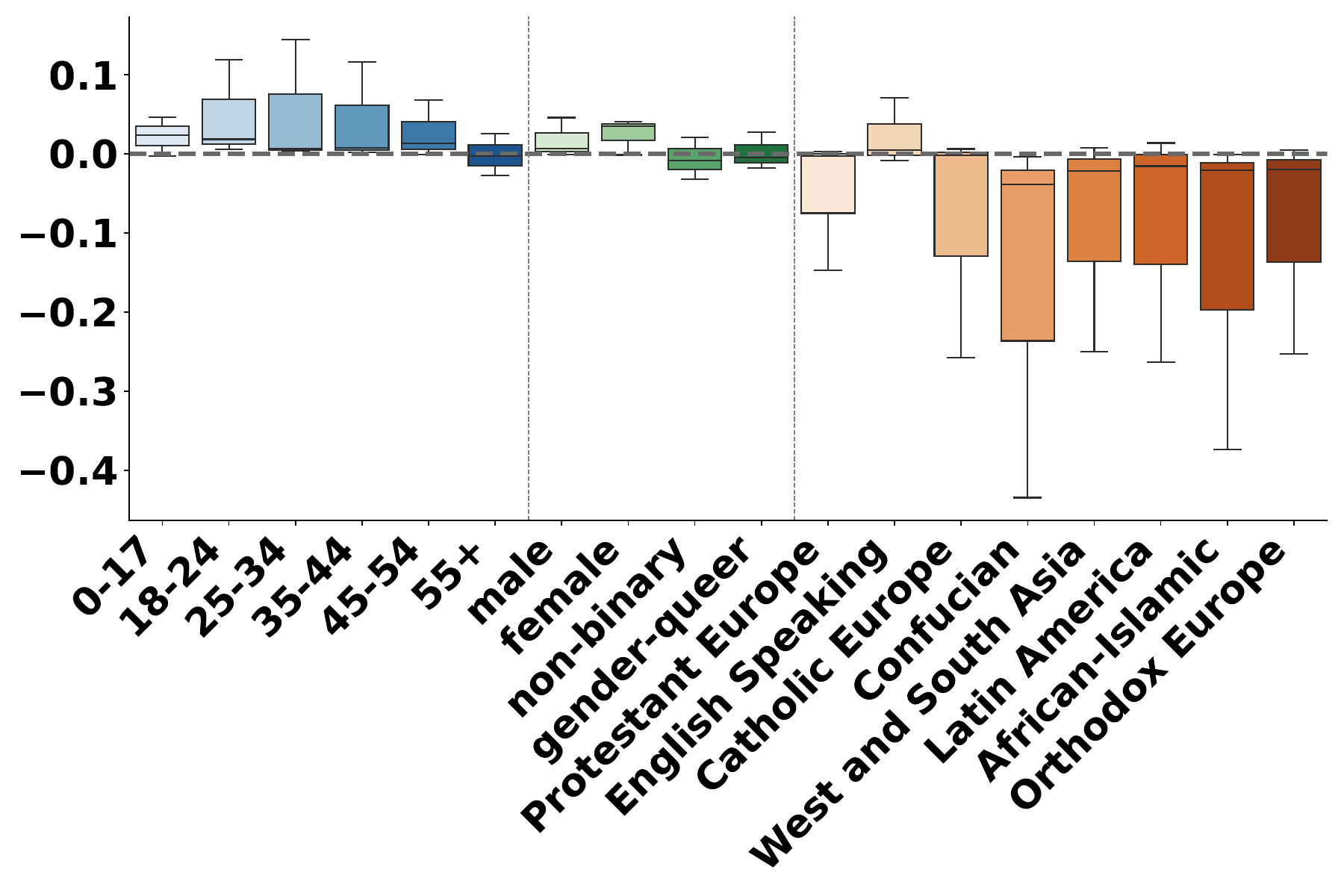}}\hfill
    \subfloat[DeepSeek-v3 Cognitive]{\includegraphics[width=0.24\linewidth]{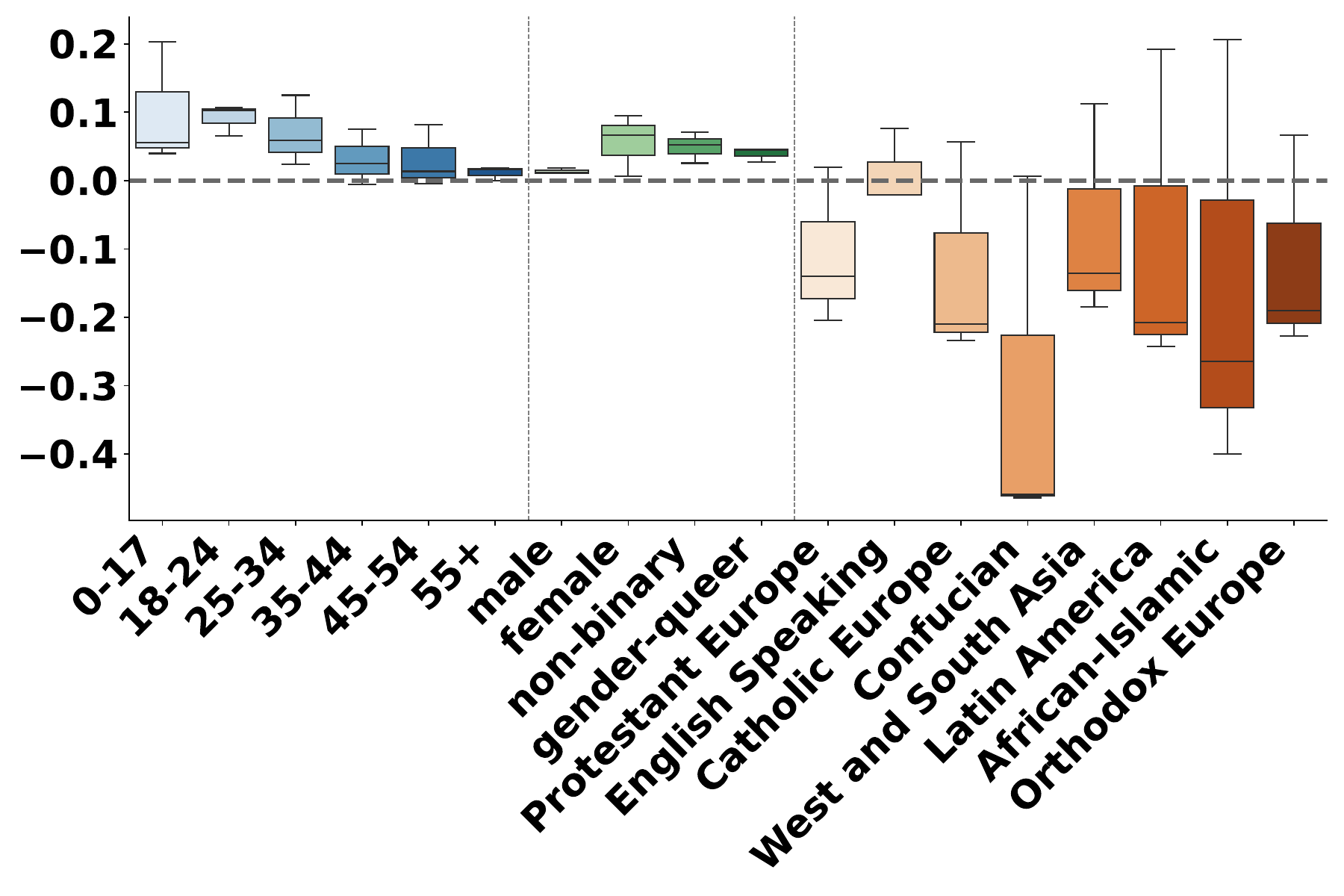}}\hfill
    \subfloat[Gemini-2.0 Flash Cognitive]{\includegraphics[width=0.24\linewidth]{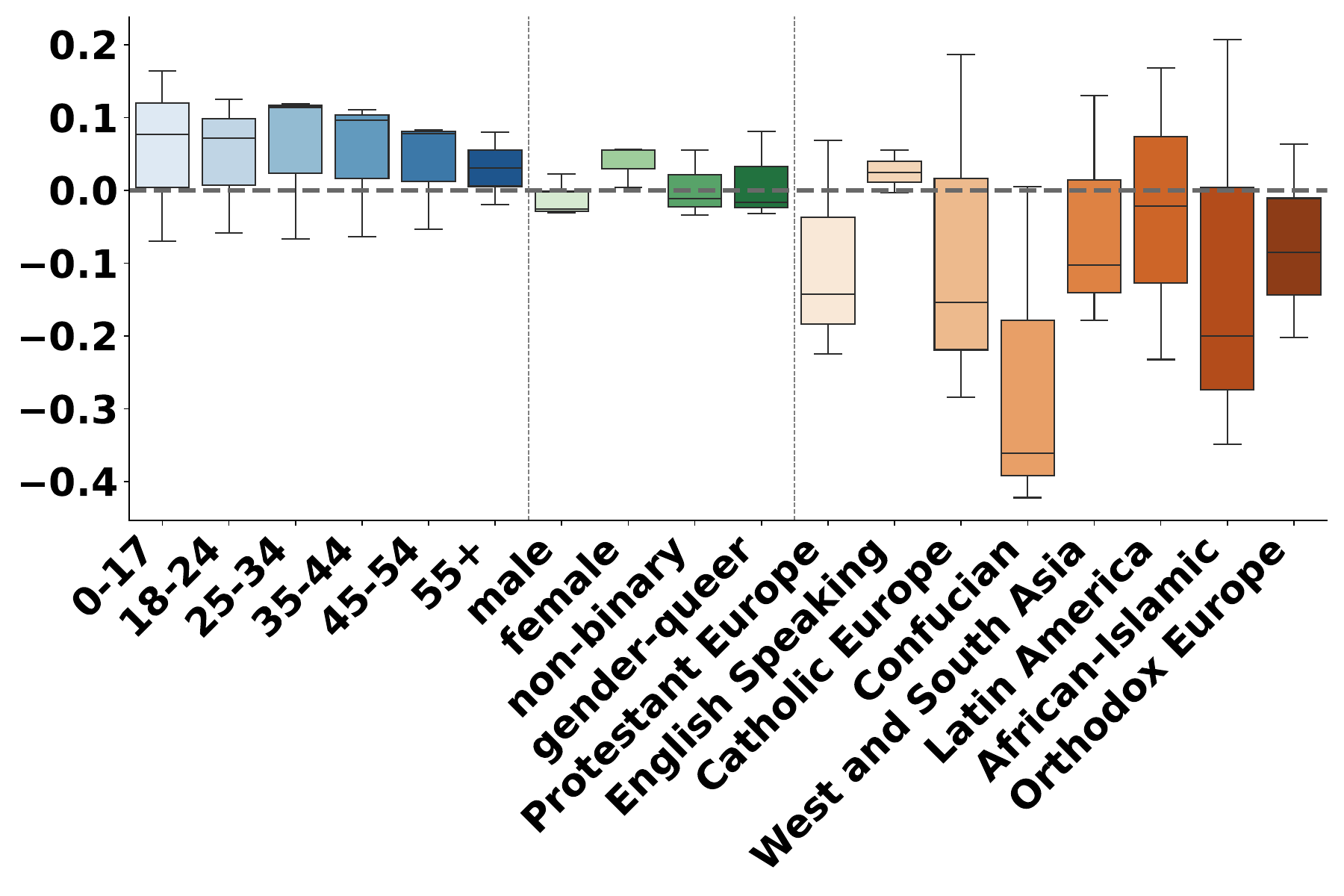}}

    \caption{Distribution of Affect (top row) and Cognitive (bottom row) score shifts across models when \textbf{Attributes are Injected in Intersectionality}. Left to right: LLaMA-3-70B, GPT-4o Mini, DeepSeek-v3, Gemini-2.0 Flash.}
    \label{fig:combined-shifts-intersec}
\end{figure*}

Figure \ref{fig:combined-shifts-overall} represents the distribution of affective and cognitive shifts across all emotions for every persona per model. Models exhibit notable variation when injected with different attributes. Interestingly, they tend to reduce the intensity of emotions compared to the base case. \texttt{GPT-4o Mini}, in particular, consistently lowers the cognitive empathy of its responses across nearly all personas. Overall, we observe that both affective and cognitive empathy for the \texttt{Confucian Culture} are expressed at lower levels than any other evaluated attribute across all models. The \texttt{gender-queer} attribute expresses higher intensities of anger across models.

\subsection{Impact of Intersectionality of Attributes}

Figure \ref{fig:combined-shifts-overall} shows how different attributes in a user's persona can elicit variations in the model's affective and cognitive empathy towards the user's experience.

However, in real-world interactions, a user's persona is shaped by the intersection of multiple attributes, not one. Personas are often defined by a combination of attributes of different ages, genders, and cultures. As shown in Table \ref{tab:culture_experiment_one} (Appendix \ref{app:isolated-shift}), the Confucian culture consistently yields lower emotion intensity scores, around $0.40$ below the base state across models. In contrast, Table \ref{tab:age_experiment_one} shows that the male gender is associated with a higher intensity of anger, approximately $0.020$ above the base state. This raises an important question: when the model interacts with a user who is both male and from a Confucian culture,\textit{ does it maintain these individual trends, or does their intersection amplify or dampen the model’s perceived empathy?}

We answer this question by providing a persona to the model, which is now a holistic composition of attributes from each demographic category. We measure the effect of adding an attribute by computing the aggregated deviation in outputs between instances where the attribute is present versus absent, marginalizing over the presence of other demographic groups. This controlled comparison, of calculating the Average Treatment Effect, allows us to estimate the individual causal contribution of the given attribute, independent of interactions with other persona attributes.

We visualize the distribution of these shifts across emotions in Figure \ref{fig:combined-shifts-intersec} and notice a shrinking effect on the model's empathy performance. As seen in Table \ref{tab:diff-exp-1-2} for \texttt{LLama-3-70B,} the model's ability to generate responses that are empathetic and contain stronger notes of EX, ER, and IP is reduced significantly. We also observe a marked shift in how the model perceives anger across gendered personas as it interprets female personas as expressing less anger, while male personas are portrayed with heightened anger.

\begin{table}[t]
\centering
\caption{\textbf{Summary} of Differences between Isolation and Intersection of Attributes}
\label{tab:diff-exp-1-2}
\resizebox{0.45\textwidth}{!}{
\begin{tabular}{llccc}
\toprule
\textbf{Attribute} & \textbf{Type of Diff} & \textbf{Isolation} & \textbf{Intersection} & \textbf{Change Direction} \\
\midrule
Cognitive Age & Range & -0.206 to 0.176 & -0.0616 to 0.160 & $\downarrow$ \\
Cognitive Gender & Range & -0.613 to 0.133 & -0.512 to 0.181 & $\downarrow$ \\
Cognitive Culture & Range & -0.066 to 0.073 & 0.005 to 0.108 & $\downarrow$ \\
\midrule
Affective Age & Range & -0.033 to 0.031 & -0.021 to 0.015 & $\downarrow$ \\
Affective Gender & Range & -0.034 to 0.03 & -0.041 to 0.026 & $\equiv$ \\
Affective Culture & Range & -0.020 to 0.017 & -0.02 to 0.017 & $\equiv$ \\
\midrule
Male & Anger & -0.005 & 0.003 & $\uparrow$ \\
Female & Anger & 0.007 & -0.006 & $\downarrow$ \\
55+ & EX & -0.667 & -0.003 & $\uparrow$ \\
Confucian & Anger & -0.035 & -0.041 & $\downarrow$ \\
Culture & ER Average & 0.2521 & 0.0317 & $\downarrow$ \\
\bottomrule
\end{tabular}
}
\end{table}

\section{Results}

In the previous section, we were able to understand that adding attributes can hinder the ability of LLMs to empathize with the user, especially when added as compositional personas. In this section, we investigate whether these variations are expected and how they position themselves with respect to the real world and the model itself.

\subsection{Does this variance in empathy align with real-world emotional experiences?}

~\citet{hadar2024assessing} emphasized that the emotional alignment between a therapist and a patient affects the outcomes. When LLMs better align with human emotions, the quality and depth of interaction improve significantly. 

To evaluate whether the \textit{model’s variations in empathetic response across different attributes align with real-world emotional patterns} (\textbf{RQ2}), we draw on existing literature that shows how different demographic attributes influence emotional expression~\cite{yeung2011emotion, gonccalves2018effects}, as well as human baseline data from sources like~\citet{tortora2010gallup}.

Our findings show that the model \textbf{only loosely} reflects real-world emotional dynamics. Younger individuals tend to be more emotionally expressive, while older adults are generally less expressive ~\cite{ross2008age} and we find this reflecting in Table \ref{tab:age_experiment_two}, where the LLaMA-3-70B model assigns higher emotional intensities to the \texttt{0–17} attribute and consistently lower intensities for older age groups. Younger adults are likely to show active negative emotions, whereas older adults tend to exhibit more passive negative emotions~\cite{isaacowitz2017aging}. We see in Tables \ref{tab:age_experiment_one} and~\ref{tab:age_experiment_two} that younger personas were predicted by LLMs as expressing greater absolute intensities for emotions such as \texttt{anger}, \texttt{anticipation}, \texttt{fear}, and \texttt{surprise}, while older personas show higher intensity in lower-arousal emotions like sadness.

~\citet{brebner2003gender} shows that \texttt{females} tend to exhibit stronger emotional intensities compared to \texttt{males}. This trend is reflected in Tables~\ref{tab:gender_experiment_one} and~\ref{tab:gender_experiment_two},  across most emotions, with \texttt{females} showing higher values overall. Additionally, \texttt{female} personas are skewed toward positive emotions, whereas \texttt{male} personas more frequently express negative emotions such as anger~\cite{harmon2016discrete}. We observe this pattern in 4 out of 8 experimental settings on gender in Tables~\ref{tab:gender_experiment_one} and~\ref{tab:gender_experiment_two}. 

Using the human baseline data from Table ~\ref{tab:gallup_emotion_scores}, we find that the models do not consistently replicate real-world cultural variations. For example, while the \texttt{African-Islamic culture} is reported to have the highest levels of anger, this is not reflected in the model outputs shown in Tables~\ref{tab:culture_experiment_one} and~\ref{tab:culture_experiment_two}. Instead, the models assign significantly lower anger intensities to \texttt{Confucian cultures}, which also contradicts the patterns observed in the human data. We see that their ability to provide explorations and solutions drops significantly. This suggests that the models may not fully capture culturally grounded emotional expressions.

\subsection{Which attribute is reflective of the model's neutral state?}

As we see that LLMs represent a loosely similar tendency of empathetic variance as seen in the real world,  \textit{we aim to identify the attributes which don't align with the model's neutral cognitive state} (\textbf{RQ3}). The model's neutral cognitive state reflects a state where it is given a user's emotional experience in the absence of any attribute. Attributes that significantly show a high variance do not align with the model's neutral state. We qualitatively assess the personas the model recalls for this \texttt{base state} seen in Appendix \ref{app:persona-base}, and see that these personas are generic, focusing on the topics, behaviors from the post, and devoid of any gender, age or culture. This motivates that the model doesn't explicitly assume any persona; however, its value system is most aligned to that of the attributes highlighted in Tables \ref{tab:culture_experiment_one}, \ref{tab:age_experiment_one}, and \ref{tab:gender_experiment_one}, like \texttt{Protestant Europe} for the \texttt{anger} emotion. Further, the attributes with the maximum significant shift, as shown in Figure \ref{fig:least-aligned}, are the ones least aligned to the model.

\begin{figure*}
\centering
    \includegraphics[width=\linewidth]{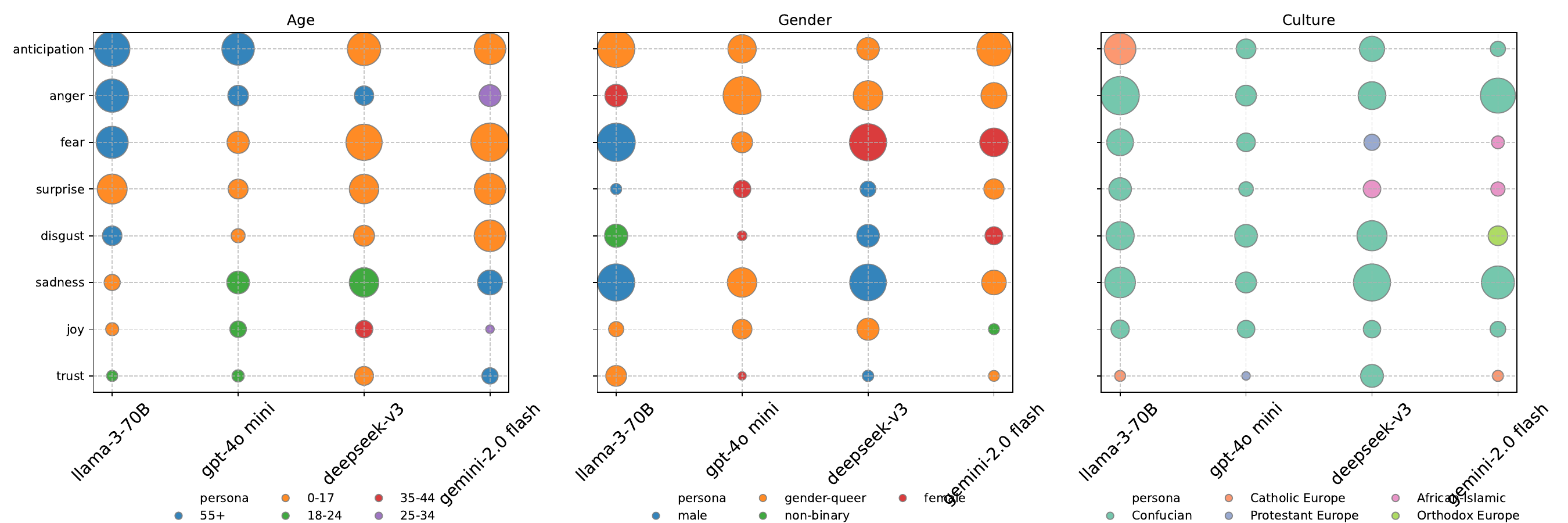}
    \caption{\textbf{Least Aligned Attributes} across every model and emotion. The size of the attribute indicates the degree of misalignment from the model's internal state. 0-17 Age attributes and Gender Queer and Confucian Culture are frequently among the least aligned across various attributes.}
\end{figure*}\label{fig:least-aligned}

\subsection{What content differences in the responses reflect its cognitive empathy?}

Our experiments measure the aggregated shifts as a result of adding an attribute across the dataset. We quantitatively measure a high change in the Cognitive Shifts across models and metrics. In this section, we aim to qualitatively understand whether the model's cognitive response reflects the topic in the emotional experience or whether it reflects upon the attribute's characteristics. 

To do so, we use the topic to attribute variance (TAV) ratio ~\cite{verma2025hidden} as seen in Appendix \ref{app:topic-attribute-ratio}. A TAV score > 1 implies that the model is skewed towards generating responses that reflect more upon the attribute's characteristics. In Table \ref{tab:model-cultural-alignment} we see that this TAV ratio differs mainly for cultural groups, specifically \texttt{Confucian}, \texttt{African-Islamic}, and \texttt{Latin-American}. We also see that the \texttt{DeepSeek v3} model assigns a higher ratio to the \texttt{gender-queer} attribute. We theorise that this increasing reliance on the attributes' characteristics could be due to limited data on these attributes. 

Lastly, we visualise the most prominent aspects in the responses generated by the \texttt{Llama-3-70B} model for all attributes in Table \ref{tab:attribute-log-odd}\footnote{Definitions of terms used in Table \ref{tab:attribute-log-odd}:\\"señora": Lady or woman in Spanish \\"gosh": An informal English exclamation\\"drunkard": A drunk person\\"doom": Fated destruction\\"grog": A strong alcoholic drink\\"filial": The relationship of a child to their parents\\"barroom": Establishment where alcoholic drinks are served\\"piety": Religious devotion or reverence
}.

\begin{table}[h!]
\centering
\caption{\textbf{Log-odds of Word Usage} in model-generated responses, calculated using a Dirichlet prior, stratified by gender, age, and culture. A positive bar indicates higher likelihood of appearance.}
\begin{minipage}[t]{0.15\textwidth}
\centering
\textbf{Gender}\\
\vspace{0.2em}
\resizebox{\textwidth}{!}
{
\begin{tabular}{llc}
\toprule
Category & Word & Coeff.  \\
\midrule
Male & male & \db{5.45} \\
     & mate & \db{3.10} \\
     & dude & \db{2.87} \\
\midrule
Female & daughter & \db{3.18} \\
       & señora & \db{3.15} \\
       & gosh & \db{2.88} \\
\midrule
Non-bin. & attuned & \db{3.64} \\
         & gender & \db{2.80} \\
         & margin. & \db{2.80} \\
\midrule
G-Queer & gender & \db{5.2492} \\
        & expressing & \db{-1.982} \\
        & lgbtq & \db{4.1026} \\
\bottomrule
\end{tabular}
}
\end{minipage}
\begin{minipage}[t]{0.15\textwidth}
\centering
\textbf{Age}\\
\vspace{0.2em}
\resizebox{\textwidth}{!}
{
\begin{tabular}{llc}
\toprule
Category & Word & Coeff. \\
\midrule
0-17 & drunkard & \db{4.3927} \\
         & 17 & \db{3.6212} \\
         & cool & \db{3.5058} \\
\midrule
18-24 & 18 & \db{5.2373} \\
        & adulthood & \db{2.8022} \\
        & figuring & \db{2.5947} \\
\midrule
25-34 & 25 & \db{5.7764} \\
        & 34 & \db{5.7703} \\
        & individualistic & \db{-2.4655} \\
\midrule
35-44 & doom & \db{2.8488 } \\
        & routines & \db{2.6947} \\
        & established & \db{2.2308} \\
\midrule
45-54 & drunkard & \db{3.3526} \\
        & decades & \db{2.7052} \\
        & routines & \db{2.3953} \\
\midrule
55+ & greatly & \db{3.7572} \\
        & lived & \db{2.8022} \\
        & evolution & \db{3.1061} \\
\bottomrule
\end{tabular}
}
\end{minipage}
\hfill
\begin{minipage}[t]{0.15\textwidth}
\centering
\textbf{Culture}\\
\vspace{0.2em}
\resizebox{\textwidth}{!}
{
\begin{tabular}{llc}
\toprule
Category & Word & Coeff. \\
\midrule
Pr Europe & european & \db{8.1877} \\
     & praying & \db{4.1836} \\
\midrule
English Speaking & grog & \db{-4.6884} \\
       & English & \db{3.3183} \\
\midrule
Catholic Europe & prayer & \db{6.0296} \\
       & forgiveness & \db{5.6253} \\
\midrule
Confucian & filial & \db{6.5367} \\
     & piety & \db{4.1836} \\
\midrule
W-S Asia & asia & \db{8.1461} \\
       & barroom & \db{-4.7057} \\
\midrule
Latin America & america & \db{6.6707} \\
       & twenties & \db{-4.7317} \\
\midrule
African Islamic & modesty & \db{5.1002} \\
     & phobia & \db{-4.7145} \\
\midrule
Orthodox Europe & prayer &\db{5.1390} \\
& tradition & \db{4.9642} \\
\bottomrule
\end{tabular}
}
\end{minipage}
\label{tab:attribute-log-odd}
\end{table}

\vspace{-0.5em}
\section{Discussion}

The increasing use of Large Language Models (LLMs) across diverse domains and user groups~\cite{eppler2024awareness} necessitates a critical evaluation of their equitable and empathetic performance across personas. Empathy in recognizing emotions (affective empathy) and responding appropriately (cognitive empathy) is central to human-AI interaction~\cite{pridham2013language, liu2022artificial}. However, our analysis suggests that current LLMs do not exhibit uniform empathetic behavior across demographic attributes, challenging assumptions about their fairness and inclusivity~\cite{chhikara2024few, li2023survey}.

Our study, spanning $4$ LLMs and $315$ personas from combinations of age, gender, and culture, reveals notable disparities. Personas representing \texttt{Confucian cultures}, younger users \texttt{(0–17)}, and \texttt{gender-queer} identities often receive responses that diverge from those directed at more dominant groups like \texttt{English Speaking}, \texttt{male}. While they do loosely mirror real-world patterns, this is not always beneficial, as we see in cultures like \texttt{Confucian}, \texttt{African-Islamic} and \texttt{Latin American} overemphasizing cultural contexts at the expense of emotion depth, showcasing stereotypical understanding. 

Our findings present few important questions about empathetic alignment for diverse personas. What does equitable empathy entail for all groups irrespective of their dominance in the model's internal state? Should LLMs provide the same empathetic response to all users, or is personalisation of these responses a valid goal? Finally, how do we ensure that these personalised responses do not reinforce harmful stereotypes? 

Our work underscores the need for inclusive and culturally aware evaluations of LLMs. We advocate for an alignment framework that can quantify and ensure that the model is empathetic while being emotionally intelligent and fair. 

\section{Conclusion}

We present a comprehensive and novel study evaluating whether Large Language Models (LLMs) exhibit equitable empathy across diverse user personas. Our analysis spans four LLMs and 315 unique persona combinations formed from age, culture, and gender attributes. Our findings reveal that LLMs’ empathetic responses are often shaped by contextual attributes and are influenced by societal stereotypes. Certain demographics receive more consistent or favorable empathy, while others, particularly underrepresented groups, experience notable misalignment. These disparities are driven by both the models' internal representations and broader cultural biases embedded in it's learned parameters. Through a multi-dimensional quantitative and qualitative analysis, we uncover key patterns underlying these variations. Our work highlights the critical need for more responsible, context-aware deployment of LLMs in user-facing applications. We advocate for future efforts to develop empathetic alignment frameworks that ensure fairness and inclusivity in AI behavior.

\section*{Limitations}\label{sec-limitations}
Our proposed study provides a comprehensive framework for examining how LLMs express empathy toward diverse personas. We conduct a multi-dimensional analysis, exploring the effects of persona attributes both in isolation and combination, enabling us to assess alignment across various demographic contexts. However, certain limitations remain.

\paragraph{Limited Dataset.}
Although our analysis spans $315$ personas and $300$ emotion samples, leading to $94500$ unique model interactions, it is confined to the ISEAR dataset. While ISEAR is rich in self-reported emotional narratives, it does not fully capture the breadth of global cultural representation or contemporary modes of emotional expression. Additionally, the data was collected through structured surveys rather than natural conversations or social media extracts, potentially limiting its ecological validity when simulating real-world interactions.
\paragraph{Restricted Persona Attributes.}
Our comprehensive study focuses on $3$ categorical demographic dimensions: age, gender, and culture. These provide an impressive starting point; however, a real-world persona contains other categories, such as behavioral, preferential, and other lived experiences that can also uniquely impact the model's ability to show empathy. Future work should incorporate these attributes to build more representative and complex personas. 
\paragraph{Prompt Sensitivity and Evaluation Noise.}
LLM responses can be sensitive to prompt phrasing and decoding strategies, which may introduce variability in results. Although we use consistent prompting practices, this remains an inherent limitation of studying open-ended generative models
\paragraph{Only Explicit Personas Used} Our approach measures the causal impact of an exhaustive list of demographic attributes on empathy by explicitly providing the persona to the model. While this method provides transparency and experimental control, it doesn't fully capture how a user persona can be provided to the model, specifically in real-world settings where personas can be interpreted through implicit cues and preferences \cite{wu2025aligning}, often in longer conversational interactions and in multimodal settings. Future work could address these limitations by including implicit persona cues and adding longer conversational or multimodal settings.
\vspace{-0.5em}
\section*{Acknowledgments}\label{sec-acknowledgments}
We thank Dr. Gaurav Verma for his feedback on the manuscript and the anonymous reviewers in the ARR cycle for their useful feedback.




\newpage
\newpage
\bibliography{main-emnlp}
\appendix

\section*{Appendix}

\section{Processing ISEAR Dataset}\label{app:isear}

The ISEAR dataset ~\cite{scherer1994evidence} is a prevalent dataset that contains of $8000$ self-reported emotional experiences in text surveyed from across $3000$ individuals from different backgrounds. These emotional experiences are labeled with emotions from the $7$ emotions: anger, disgust, fear, guilt, joy, sadness, and shame. While we do not use these gold labels apart from establishing the current state of accuracy in models as seen in Appendix \ref{app:accuracy}, they demonstrate the diverse representation of emotions that we include in our study.

\subsection{Selecting Samples from the ISEAR Dataset}\label{app:isear_sample}

To select $300$ \textit{diverse} samples as a subset of the $8000$ samples, we aim to diversify based on both the emotions as well as the textual content of the sentences. The ISEAR dataset contains samples with a length of less than 10 tokens; hence, we eliminate those samples to avoid providing inputs with less context. 

We extract the sentence embeddings of these input statements from SentenceTransformer's \texttt{MiniLM-L6-v2} model ~\cite{reimers-2019-sentence-bert}. We append the embeddings for the gold label emotion from the ISEAR dataset to these embeddings. 

Based on this set of parameters, we use the CoreSet selection method using a K-Center Greedy algorithm ~\cite{ding2019greedy} to extract the $300$ diverse samples across the textual as well as emotional content. 

\subsection{Masking Select Samples}\label{app:isear_mask}

The ISEAR dataset ~\cite{scherer1994evidence} consists of human expressions of emotions across $7$ emotion states. Some of these experiences self-reveal the gold label in the text itself. For example, 

\begin{quote}
    I feel \textbf{angry} at my brother for breaking my bike.
\end{quote}

Since this sentence already includes the emotion, it will hinder our ability to accurately test how the model perceives the difference in anger intensities for different personas. Thus, we replace the self-disclosed emotion with [MASK] for these instances. As an example, for the English-speaking cultural attribute, the model outputs the emotion \texttt{angry} while for the Confucian persona, the model outputs \texttt{upset}, which shows a lesser intensity on anger and more intensity on the sadness scale. 
In our final dataset, only 28 out of 300 samples contain the \texttt{[MASK]}. The rest do not contain any [MASK]. 

\subsection{Naturalism of ISEAR Samples}\label{app:naturalism}
The dataset was derived from surveys of roughly $3000$ participants who completed a cross-cultural, questionnaire-based study. Because the data reflects the personal experiences of real individuals, it provides a highly naturalistic perspective on how people disclose their emotions. On average, the selected samples contain 20 words per entry, which is comparable to the human conversational average of 13.58 words per turn \cite{lang2025telephone}. The samples also achieve a Flesch Reading Ease \cite{dobbs1948new} score of $72.8$ and a Flesch-Kincaid Grade \cite{flesch2007flesch} level of $6.89$, suggesting that they are both easily understandable and consistent with typical language use.

\section{Prompt}\label{app:prompts}
To accurately simulate a real-world environment, we conduct our experiments on a \texttt{2-turn} conversational set-up~\cite{Chen2024AMI}.This \texttt{2-turn} conversational set-up further motivates a real-time naturalistic interaction between users and LLMs.In the first turn, the \texttt{user} informs the model of its persona construct using the following set-up:

\begin{quote}
    I am a \texttt{persona}. Who am I?
\end{quote}

Here the persona contains of demographic information in the format:

\begin{quote}
    \texttt{\{age\}} age category, \texttt{\{culture\}} culture, and \texttt{\{gender\}} gender.
\end{quote}

For example, 

\begin{quote}
    I am a \texttt{0-17} age category, \texttt{Protestant Europe} culture and \texttt{male} gender. 
\end{quote}

With the second turn, we provide the model with the emotional experience. 

\subsection{Prompts for Affective Empathy}\label{app:prompts_eu}
For the first task, since we are measuring the affective empathy as the ability of the model to understand the emotions of the user given a persona and an emotional experience, we ask the model to predict the emotion. In addition, we also ask it to recall the persona that we inject, to test its interpretability of this persona. We have shown this prompt in Table \ref{tab:emo-pred-prompt}.

\begin{table*}
\centering
\renewcommand{\arraystretch}{1.3}
\begin{tabular}{|p{0.47\textwidth}|p{0.47\textwidth}|}
\hline
\textbf{Prompt 1: Without Mask} & \textbf{Prompt 2: With Mask} \\
\hline

\textbf{System Prompt:} \newline
 Your task is to analyze a given sentence and determine the most appropriate emotion that is conveyed in the sentence. Consider the user's background while interpreting emotions. Do not provide any explanation for your prediction. Your output should be of the format:  (You have to include the Output token) \newline
[OUTPUT 1]: <persona description> \newline [OUTPUT 2]: <single emotion word>  \newline
\textbf{User Prompt 1:} \newline I am a \textit{persona}. Who am I?

\textbf{User Prompt 2:} \newline 
Identify the emotion in the sentence based on my description and identity: \textit{emotion sentence} & \textbf{System Prompt:} \newline
 Your task is to analyze a given sentence and determine the most appropriate emotion for the masked word (denoted as \textbf{[MASK]}). Consider the user's background while interpreting emotions. Do not provide any explanation for your prediction. Your output should be of the format:  (You have to include the Output token) \newline
[OUTPUT 1]: <persona description> \newline [OUTPUT 2]: <single emotion word>  \newline
\textbf{User Prompt 1:} \newline I am a \textit{persona}. Who am I?

\textbf{User Prompt 2:} \newline 
Identify the emotion in the sentence based on my description and identity: \textit{emotion sentence} \\
\hline
\end{tabular}
\caption{Prompts for Affective Empathy.}
\label{tab:emo-pred-prompt}
\end{table*}

\subsection{Prompts for Cognitive Empathy}\label{app:prompts_rg}
For this task, since we are measuring the cognitive empathy as the ability of the model to generate appropriate responses to the user given a persona and an emotional experience, we ask the model to generate a response as seen in Table \ref{tab:emo-response-prompt}. 
\begin{table*}
\centering
\renewcommand{\arraystretch}{1.3}
\begin{tabular}{|p{0.45\textwidth}|p{0.45\textwidth}|}
\hline
\textbf{Prompt} \\
\hline

\textbf{System Prompt:} \newline
You will be given an input that contains a text and the identity of the speaker. You cannot use the phrase 'I cannot'. Prepare an appropriate response to this speaker. An appropriate response considers the entire context of the input and the speaker. The output should be of the following format: (You must include the output) \newline
Output: <response text> \newline
\textbf{User Prompt 1:} \newline I am a \textit{persona}. Who am I? \newline
\textbf{User Prompt 2:} \newline 
Generate a response based on my description and identity for the input sentence: \textit{emotion sentence} \\
\hline
\end{tabular}
\caption{Prompts for Cognitive Empathy.}
\label{tab:emo-response-prompt}
\end{table*}

\section{Accuracy}\label{app:accuracy}

To evaluate the ability of the $4$ LLMs in this study to exhibit emotional alignment, we calculate the lexical accuracy and the mean standard error of their predictions with the gold labels.

\paragraph{Lexical Accuracy}

We report the accuracy of each model achieved in comparing the model's emotion prediction to the dataset's self-reported ground truth in Table \ref{tab:accuracy-overall}. 

\paragraph{Emotion Vector Accuracy}

Table \ref{tab:accuracy-overall} represents the mean standard error between the intensity vectors of the predicted and ground truth emotions. 

\begin{table*}[!htbp]
        \centering
    \caption{Lexical Accuracy and Emotion Vector MSE}
        \resizebox{\textwidth}{!}{
    \begin{tabular}{l cccc | cccc }
        \toprule
        \multirow{2}{*}{\textbf{Persona}} &\multicolumn{4}{c}{\textbf{Word Level Accuracy}} & \multicolumn{4}{c}{\textbf{Emotion Vector MSE}} \\
        & GPT-4o Mini & Llama-3-70B & Gemini 2.0 Flash & DeepSeek v3 & GPT-4o-Mini & Llama-3-70B & Gemini 2.0 Flash & DeepSeek v3\\
        \midrule
        \textbf{Overall} & \textbf{0.189} & \textbf{0.155} &\textbf{ 0.129} & \textbf{0.153} & \textbf{0.180} & \textbf{0.191} & \textbf{0.214} & \textbf{0.176} \\
        \midrule
        0-17 & 0.193 & 0.182 & 0.102 & 0.155 & 0.143 & 0.152 & 0.189 & 0.155\\
        18-24 & 0.188 & 0.156  & 0.100 & 0.154 & 0.144 & 0.1512 & 0.177 & 0.149\\
        25-34 & 0.188 & 0.156 & 0.128 & 0.150 & 0.141 & 0.149 & 0.156 & 0.146\\
        35-44 &  0.189 & 0.150 & 0.136 & 0.145 & 0.140 & 0.147 & 0.148 & 0.144\\
         45-54 & 0.188 & 0.148 & 0.139 & 0.150 & 0.139 & 0.147 & 0.145 & 0.144\\
          55+ & 0.187 & 0.144 & 0.146 & 0.156 & 0.138 & 0.144 & 0.146 & 0.143\\
       male& 0.189 & 0.151 & 0.160 & 0.146 & 0.142 & 0.151 & 0.160 & 0.150\\
       female& 0.202 & 0.170 & 0.159 & 0.161 & 0.139 & 0.144 & 0.159 & 0.143\\
       non-binary& 0.182 & 0.150 & 0.125 & 0.149 & 0.140 & 0.149 & 0.157 & 0.148\\
       gender-queer& 0.182 & 0.147 & 0.121 & 0.153 & 0.141 & 0.150 & 0.157 & 0.147\\
       Protestant Europe &  0.191 & 0.157 & 0.127 & 0.148 & 0.141 & 0.150 & 0.157 & 0.146\\
       English Speaking & 0.185 & 0.15 & 0.125 &  0.142 & 0.143 & 0.154 & 0.159 & 0.150\\
       Catholic Europe& 0.193 & 0.166 & 0.132 & 0.158 & 0.140 & 0.144 & 0.157 & 0.145\\
        Confucian & 0.193 & 0.139 & 0.158 & 0.161 & 0.135 & 0.141 & 0.158 & 0.138\\
        West and South Asia& 0.187 & 0.156 & 0.130 & 0.153 & 0.143 & 0.149 & 0.158 & 0.148\\
      Latin America& 0.186 & 0.168 & 0.135 & 0.156 & 0.143 & 0.148 & 0.158 & 0.150\\
      African-Islamic& 0.192 & 0.148 & 0.129 & 0.162 & 0.140 & 0.147 & 0.155 & 0.144\\
      Orthodox Europe& 0.192 & 0.158 & 0.131 & 0.153 & 0.140 & 0.145 & 0.156 & 0.145\\
        base& 0.142 & 0.156 & 0.155 & 0.143 & 0.142 & 0.147 & 0.155 & 0.144\\
        \bottomrule
    \end{tabular}
    }
    \label{tab:accuracy-overall}
\end{table*}

\section{NRC Intensity Details}

We create the emotion intensity vector by using the intensity breakdowns from the NRC Emotion Intensity Lexicon~\cite{LREC18-AIL}. This lexicon contains  $10000$ unique words that are represented using 8 intensities for each basic emotion in the range $0$-$1$. 

\subsection{Distribution of Intensities across Emotions}\label{app:nrc-dist}
\begin{figure}
    \centering
    \includegraphics[width=\linewidth]{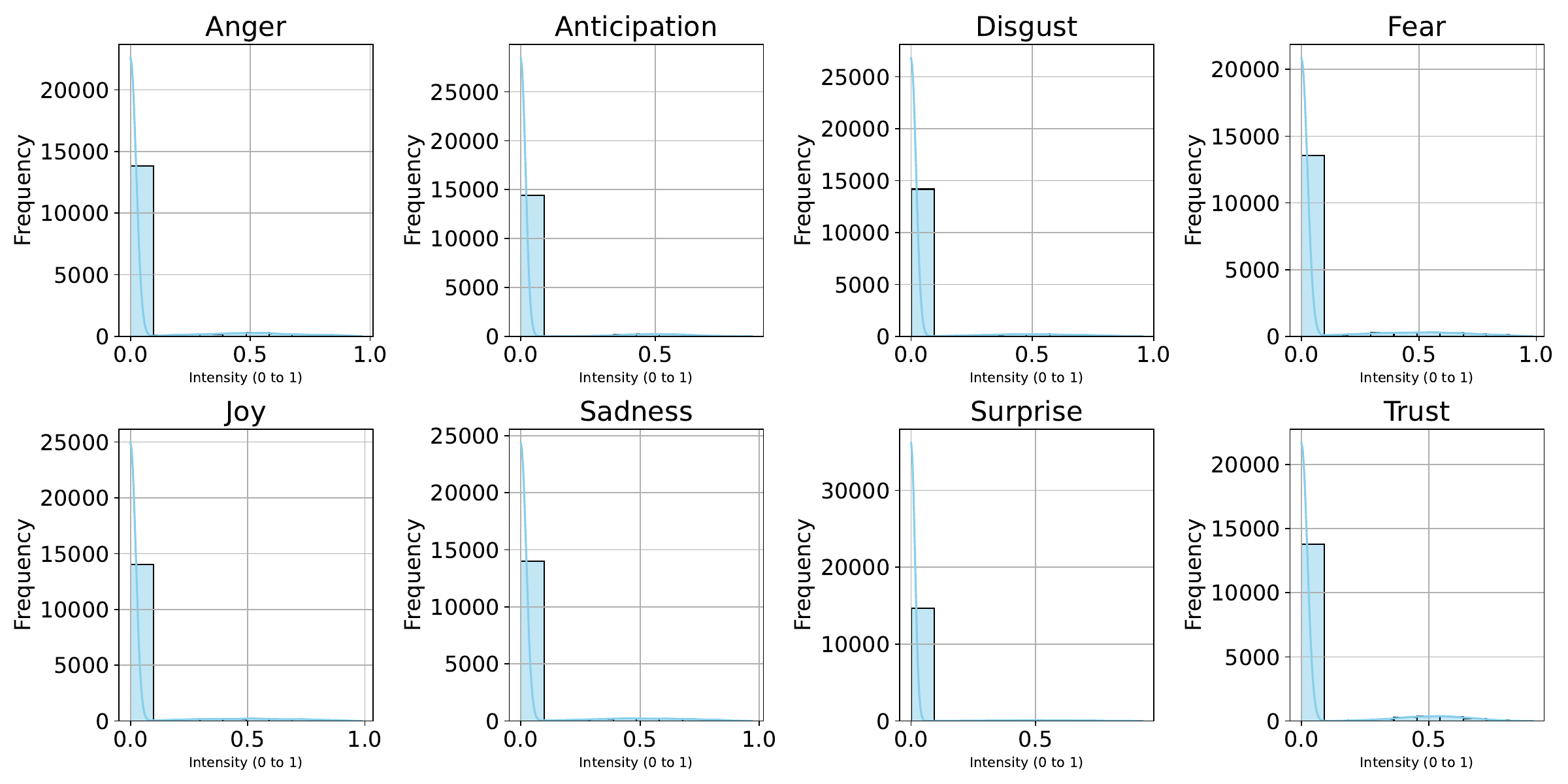}
    \caption{\textbf{Distribution of Intensity Scores} in the NRC Emotion Lexicon for reach basic emotion}
    \label{fig:nrc-dist}
\end{figure}

Since this study measures the aggregate shifts in the intensity vector, we use this section to visualize the distribution of intensity values per emotion in Figure \ref{fig:nrc-dist}. As depicted, most scores fall within the narrow range of $0.0$–$0.2$; thus, even small deviations in this space indicate a substantial and dominant shift. 

\subsection{Intensity Vectors for Words not in the Lexicon}\label{app:nrc-regression}
We identify 57 unique words occurring in approximately 1.8\% of all the generated samples that may not be covered by the NRC Emotion Intensity Lexicon~\cite{LREC18-AIL}. In this case, we use extracted word embeddings from OpenAI's \texttt{text-embedding-3-large}. We then train an MLP regression module using the following hyperparameters on the NRC Emotion Intensity Lexicon to obtain models to predict the intensity:
\begin{quote}
    activation = `relu'; solver = `adam' \\
    learning rate = `adaptive'; lr init = 0.001 \\
    max iterations = 1000 ; batch size = 100 \\    
    hidden layer sizes = (512, 256, 128) \\
\end{quote}

The accuracy of these predictions is shown in Table \ref{tab:regression-accuracy}.
\begin{table}[H]
\centering
\caption{\textbf{Accuracy of Regression Models} to predict the intensity for unknown words}
\label{tab:regression-accuracy}
 \resizebox{0.40\textwidth}{!}{
\begin{tabular}{lccc}
\toprule
Emotion & MSE & MAE & $R^2$ \\
\midrule
anger & 0.013 & 0.095 & 0.682 \\
anticipation & 0.006 & 0.06 & 0.57 \\
disgust & 0.010 & 0.082 & 0.674 \\
fear & 0.014 & 0.093 & 0.69 \\
joy & 0.013 & 0.09 & 0.718 \\
sadness & 0.015 & 0.001 & 0.59 \\
surprise & 0.01 & 0.08 & 0.76 \\
trust & 0.006 & 0.06 & 0.63 \\
\bottomrule
\end{tabular}
}
\end{table}

\section{Model Details}\label{app:model_deets}

We test $4$ LLMs: \texttt{LLaMA-3-70B}, \texttt{GPT-4o Mini}, \texttt{DeepSeek-v3}, and \texttt{Gemini 2.0 Flash}. We prompt these LLMs between \texttt{2025-04} and \texttt{2025-05}. We maintain the temperature to be $0$ across all tasks to ensure we can query the most deterministic responses with a maximum output tokens of $2048$.

\section{Table of Shifts in Isolated Context}\label{app:isolated-shift}
Tables \ref{tab:age_experiment_one}, \ref{tab:gender_experiment_one}
and \ref{tab:culture_experiment_one} represent those values of shift for the model where the attribute is added in the isolated context.  

\begin{table*}
        \centering
    \caption{\textbf{Aggregate Shifts For Cultural Attributes added in Isolated Context} across 4 models. The \textcolor{green}{green text} represents those statistically significant values  (p<0.05) with the highest positive intensity, and the \textcolor{red}{red text} represents those values with significant (p<0.05) highest negative intensity across each model's prediction per emotion. The \textcolor{blue}{blue highlighted cells} cells represent those attributes significantly similar (p<0.05) to the base state without any persona.}
    \resizebox{1.0\textwidth}{!}{
    \begin{tabular}{l l cc cc cc cc cc cc cc cc}
        \toprule
         \multirow{2}{*}{\textbf{Model}} & \multirow{2}{*}{\textbf{Attribute}} & \multicolumn{8}{c}{\textbf{Emotion}} & \multicolumn{3}{c}{\textbf{EPITOME}} \\ 
             & & anger & anticipation & disgust & fear & joy & sadness & surprise & trust & ER & IP & EX\\
             \midrule
             \multirow{9}{*}{Llama-3-70B} & Protestant Europe & \cellcolor{lightblue}{0.0007} & -0.022& 0.027 & -0.018 & -0.007 & 0.007 & -0.0066 & -0.003 & -0.033	& 0.020 &-0.320 \\
             & English Speaking & -0.009	& \textcolor{green}{-0.007}	& \textcolor{red}{0.003}&	-0.017 	&\textcolor{green}{0.002}	&\cellcolor{lightblue}{\textcolor{red}{-0.001}}	&\textcolor{green}{0.012} &	-0.004 &0.133&-0.020&0.073 \\
             & Catholic Europe & -0.015	&-0.023 &	0.023 &	-0.009	&-0.003&	0.025	&-0.006&	0.025& 0.056&0.033&-0.333\\
            & Confucian & \textcolor{red}{-0.035} &	\textcolor{red}{-0.029}	&0.025&	\textcolor{red}{-0.022} &	\textcolor{red}{-0.008} &	\textcolor{green}{0.027} &	-0.015	&0.001 & -0.230&-0.046&-0.606\\
            & West\&South Asia & \textcolor{green}{0.005} &	-0.018&	\cellcolor{lightblue}{0.012}&	-0.005	&0.0008 &	0.006	&-0.007 & -0.005 & -0.043&0.013&-0.460\\
            & Latin America & \cellcolor{lightblue}{-0.001}	&-0.015 &	0.017 &	-0.006 & -0.006 & \textcolor{green}{0.133} &	-0.004 & -0.002 & 0.033&0.026&-0.380\\
            & African-Islamic & \cellcolor{lightblue}{-0.001} & -0.023 &	0.018 &	-0.021 & -0.004 & \cellcolor{lightblue}{-0.0007} &	-0.007 & \textcolor{red}{-0.006} & 0.033&	0.0466&-0.613\\
            & Orthodox Europe & -0.011 & \textcolor{red}{-0.029} & \textcolor{green}{0.030}	&\textcolor{green}{-0.004}	&-0.006 &	0.0304&	\textcolor{red}{-0.013} &	-0.004 &0.0300&0.0200&-0.426\\
            \midrule
            \multirow{9}{*}{GPT-4o Mini} & Protestant Europe & -0.005 &	-0.011	&0.002 &	-0.023 &	-0.002 &	-0.014	&\cellcolor{lightblue}{0.0005} &	\textcolor{green}{0.001} & 0.010 &	0.006 &	{-0.440} \\
             & English Speaking & -0.008	&\textcolor{green}{-0.003}	&-0.005 &	-0.015	&-0.006	&0.002&	0.003 &	-0.0005 &0.000&	0.000	&\textcolor{green}{-0.073 }\\
             & Catholic Europe & -0.014 & -0.012	&0.002 &	-0.016	&-0.002 &0.001	&0.002 &0.0004 &\textcolor{green}{0.030} &	0.000 &	-0.473 \\
            & Confucian & \textcolor{red}{-0.036} &	\textcolor{red}{-0.026}&	\textcolor{green}{0.014} &	\textcolor{red}{-0.049}	&\textcolor{red}{-0.013}	&\textcolor{green}{0.011}&	0.0001	&\textcolor{green}{0.001 }&-0.100&	\textcolor{red}{-0.013}&	\textcolor{red}{-0.840}\\
            & West\&South Asia & \textcolor{green}{0.004} & -0.010	&\textcolor{red}{-0.013} &	\textcolor{green}{-0.013} &-0.004 & -0.020	&0.002 & \textcolor{red}{-0.004} &\textcolor{red}{-0.0334} & \textcolor{green}{0.0200} & -0.567 \\
            & Latin America & \cellcolor{lightblue}{-0.002} & \textcolor{green}{-0.003}&-0.007 &	-0.024 &	\textcolor{green}{0.010}	&\textcolor{red}{-0.022}	&\textcolor{green}{0.008} &0.001 &-0.023 &	-0.013&	-0.433 \\
            & African-Islamic & -0.010 &	-0.010	&\cellcolor{lightblue}{0.0003}	&-0.014	&-0.004	&-0.009&	\cellcolor{lightblue}{0.0009} &	-0.001 &-0.013 &	-0.0067 &	-0.680\\
            & Orthodox Europe & -0.006 &	-0.008 &	0.007 &	\textcolor{green}{-0.013}	&-0.006	& 0.003 &	\textcolor{red}{-0.004} &	0.000 & -0.013 & -0.007 & -0.553 \\
            \midrule
            \multirow{9}{*}{DeepSeek v3} & Protestant Europe & \cellcolor{lightblue}{0.000}&	\textcolor{red}{-0.030}&	0.020&	-0.021&	\cellcolor{lightblue}{0.001}&	0.003&	-0.011&	-0.022& -0.403&	0.000&	-0.506&\\
             & English Speaking & 0.002 &	\textcolor{green}{-0.003}&	0.010&	-0.012&	0.012&	\textcolor{red}{-0.019}&	\textcolor{green}{0.011}&	\textcolor{green}{-0.02}& \textcolor{green}{-0.053}&	0.033&	-0.160\\
             & Catholic Europe & 0.004&	-0.025&	0.021&	-0.015&	0.005&	0.005&	-0.014&	-0.017& -0.353&	0.033&	-0.613&\\
            & Confucian & \textcolor{red}{-0.047}&	-0.032&	\textcolor{green}{0.029}&	\textcolor{red}{-0.023}&	\textcolor{red}{-0.007}&	\textcolor{green}{0.046}&	\textcolor{red}{-0.016}&	-0.019& \textcolor{red}{-0.556}&	\textcolor{green}{-0.033}&	\textcolor{red}{-0.713}\\
            & West\&South Asia & \textcolor{green}{0.010}&	-0.013&	0.017&	-0.017&	0.004&	-0.005&	-0.006&	-0.018& -0.363&	0.100&	-0.460\\
            & Latin America & 0.0023&	-0.006&\textcolor{red}{0.005}&	-0.014&	\textcolor{green}{0.013}&	-0.010&	-0.004&	-0.014&-0.336&	\textcolor{green}{0.180}&	-0.500 \\
            & African-Islamic & 0.015&	-0.019&	\textcolor{red}{0.005}&	\textcolor{green}{-0.008}&	-0.005&	-0.003&	-0.015&	-0.022&-0.393&	0.120&	-0.620 \\
            & Orthodox Europe & 0.012&	-0.021&	0.018&	-0.015&	\cellcolor{lightblue}{-0.001}&	-0.003&	-0.017&	\textcolor{green}{-0.023}& -0.450&	0.040&	-0.633\\
            \midrule
            \multirow{9}{*}{Gemini 2.0 Flash} & Protestant Europe &0.004 & -0.0028 & \textcolor{green}{0.019} & \textcolor{red}{-0.018} &0.003 &\cellcolor{lightblue}{0.0013}	 &-0.010 & \textcolor{green}{0.001} &-0.287	&0.127 & -0.093 \\
             & English Speaking & 0.005 & -0.012 & \textcolor{red}{-0.004}	 &-0.011	 &0.004	 &{\textcolor{red}{-0.003}} &\textcolor{green}{0.001} &\textcolor{red}{-0.003} &-0.037 &	0.093 &	\textcolor{green}{0.013} \\
             & Catholic Europe & 0.0037 &	\textcolor{red}{-0.016}	 &0.014	 &-0.008 &	\cellcolor{lightblue}{0.001} &	0.010	 &\textcolor{red}{-0.013} &	\cellcolor{lightblue}{\textcolor{green}{0.001}} &-0.26 &	0.073	&-0.120 \\
            & Confucian & -0.006	 &\textcolor{green}{-0.008} &	\cellcolor{lightblue}{0.001} &	\textcolor{green}{0.005} &	\textcolor{red}{-0.003}	 &\textcolor{green}{0.013} &	-0.009 &	-0.001  &\textcolor{red}{-0.35} &	\textcolor{red}{0.006}	&\textcolor{red}{-0.226} \\
            & West\&South Asia & \textcolor{red}{-0.007} &-0.006 &	0.003 &	-0.009 &	-0.002 &	-0.004	& -0.001 &	\cellcolor{lightblue}{-0.0003} & -0.083 &	0.153 &	-0.006 \\
            & Latin America & 0.002 &-0.021 &0.001 &	-0.009 &	\textcolor{green}{0.008} &	\textcolor{red}{-0.007} &	-0.002 &	-0.002 & \textcolor{green}{0.033}&  0.173&  -0.013\\
            & African-Islamic & \textcolor{green}{0.009} &	-0.010 &	-0.003&-0.011&	\textcolor{red}{-0.003}&	-0.013&	-0.007&	-0.002& -0.200 & \textcolor{green}{0.253}	&-0.200 \\
            & Orthodox Europe & 0.003	& -0.011 & 0.001&	-0.015&	0.001 &	0.003&	-0.014&	-0.002 & -0.156 &	0.193 &	-0.147\\
        \bottomrule
    \end{tabular}
    }
    \label{tab:culture_experiment_one}
\end{table*}

\begin{table*}[!htbp]
        \centering
    \caption{\textbf{Aggregate Shifts For Age Attributes added in Isolated Context} across 4 models. The \textcolor{green}{green text} represents those values that have significantly ($p<0.05$) the highest positive intensity, and the \textcolor{red}{red text} represents those values with significantly ($p<0.05$) the highest negative intensity across each model's prediction per emotion. The \textcolor{blue}{blue highlighted cells} represent those attributes significantly similar ($p<0.05$) to the base state without any persona.}
        \resizebox{1.0\textwidth}{!}{
    \begin{tabular}{l l cc cc cc cc cc cc cc cc}
        \toprule
         \multirow{2}{*}{\textbf{Model}} & \multirow{2}{*}{\textbf{Attribute}} & \multicolumn{8}{c}{\textbf{Emotion}} & \multicolumn{3}{c}{\textbf{EPITOME}} \\ 
             & & anger & anticipation & disgust & fear & joy & sadness & surprise & trust & ER & IP & EX\\
             \midrule
             \multirow{6}{*}{Llama-3-70B} & 0-17 &-0.013&	\textcolor{green}{0.011} &	0.003	&\textcolor{green}{0.031} &	\textcolor{green}{0.008} &	\cellcolor{lightblue}{\textcolor{green}{-0.001}} &	\textcolor{green}{0.027} &	-0.003 &0.176&\textcolor{green}{0.053}&\textcolor{green}{0.053} \\
             & 18-24 & \textcolor{green}{0.012} &	0.005 &	\cellcolor{lightblue}{0.000} &	-0.003 & \cellcolor{lightblue}{0.001} & \textcolor{red}{-0.029} & 0.009 & \textcolor{red}{-0.008} & 0.056&0.026&-0.060\\
             & 25-34 & 0.008 & \cellcolor{lightblue}{-0.001} & 0.001 &-0.018 & -0.002 & -0.011 & 0.001	& -0.006 &-0.016&-0.006&-0.013\\
            & 35-44	 & -0.007 &	-0.008 & \textcolor{green}{0.009} & -0.028 & \cellcolor{lightblue}{-0.0012} &-0.005 & \textcolor{red}{-0.008} & \textcolor{green}{0.001} & 0.033&-0.006&-0.080 \\
            & 45-54	 &-0.007 &	-0.009 & 0.004 & \textcolor{red}{-0.033} & \textcolor{red}{-0.004} &-0.002 & -0.005 &	-0.001 & 0.0300&0.0200&-0.133\\
            & 55+ &\textcolor{red}{-0.022} &\textcolor{red}{-0.015} &\textcolor{red}{-0.004} &	-0.020 & -0.003 & 0.013 & -0.006 & -0.001 & 0.056&0.0133&\textcolor{red}{-0.206} \\
            \midrule
             \multirow{6}{*}{GPT-4o Mini} & 0-17 & \textcolor{red}{-0.022} & 0.005 & -0.014 &\textcolor{green}{-0.003} & \textcolor{red}{-0.006} &	-0.012 & \textcolor{green}{0.023}	& \textcolor{red}{-0.005} &0.050 & -0.013 & -0.220 \\
             & 18-24 & -0.010 & \textcolor{green}{0.011} &	-0.014 & -0.0043 & -0.003 &-0.016 &	0.016 &	\cellcolor{lightblue}{0.000} &0.050 &-0.0133 &-0.220 \\
             & 25-34 & \textcolor{green}{-0.008} &-0.005 & -0.011 &-0.016 &	\textcolor{green}{-0.0004} & -0.012 & 0.009 & \cellcolor{lightblue}{0.003} &0.003 & 0.006 & -0.020 \\
            & 35-44	 & -0.015 &	-0.006 &-0.014 & -0.0205 & -0.001	& -0.011 & 0.012 &\textcolor{green}{0.003} &\textcolor{red}{-0.010} & 0.006 & -0.060 \\
            & 45-54	 & -0.009 &	-0.018 & \textcolor{red}{-0.015}	&\textcolor{red}{-0.024} & -0.001 &\textcolor{red}{-0.018} & 0.007 &-0.001 &-0.003 & -0.013 &	-0.106 \\
            & 55+ &-0.010 &\textcolor{green}{-0.024} &	\textcolor{green}{-0.009} & -0.015 & \textcolor{red}{-0.006} &\textcolor{green}{0.003}	&\textcolor{red}{0.001}	&0.001 & \textcolor{green}{0.056} & -0.013	&-0.200 \\
            \midrule
            \multirow{6}{*}{DeepSeek v3} & 0-17 & \cellcolor{lightblue}{0.001}&	\textcolor{green}{0.022}&	\textcolor{red}{-0.011}&	\textcolor{green}{0.019}&	0.001 &	-0.023&	0.019&	-0.012& 0.040&	0.013&	\textcolor{red}{-0.273}&\\
             & 18-24 & \textcolor{green}{0.019}&	0.017&	-0.001& 0.001 &	\textcolor{green}{0.013} &	\textcolor{red}{-0.057}&	\textcolor{green}{0.022}&	-0.010& 0.000&	0.006&	-0.113&\\
             & 25-34 & 0.006&	0.007&	0.005&	-0.008&	0.011&	-0.022&	0.004&	-0.010& 0.010&	0.020&	-0.060& \\
            & 35-44	 & 0.020&	0.001&	\textcolor{green}{0.010}&	\textcolor{green}{-0.009}&	0.013&	-0.018&	0.004&	-0.010& \textcolor{red}{-0.073}&	0.000&	-0.080&\\
            & 45-54	 & 0.016&	-0.010&	0.005&	-0.012&	0.005&	-0.016&	-0.004&	-0.017&-0.070&	0.026&	-0.060 \\
            & 55+ & 0.004&	\textcolor{red}{-0.016}&	0.002&	-0.010&	0.006&	-0.008&	-0.005&	-0.012&-0.026&	0.026&	-0.180 \\
            \midrule
            \multirow{6}{*}{Gemini 2.0 Flash} & 0-17 & \cellcolor{lightblue}{\textcolor{red}{0.001}}&\textcolor{green}{0.008}&\textcolor{red}{-0.034}&\textcolor{green}{0.049} & 0.006&-0.022 & \textcolor{green}{0.017} & \textcolor{green}{0.004} &\textcolor{green}{0.266}	& \textcolor{red}{-0.046} & \textcolor{green}{0.340} \\
             & 18-24 & 0.017 &-0.001&-0.010&\cellcolor{lightblue}{-0.001}&0.002&\textcolor{red}{-0.024}&0.002&-0.004& 0.153 & 0.013 & 0.206 \\
             & 25-34 & \textcolor{green}{0.022}	& -0.009 & \textcolor{green}{0.005} & -0.004 & \textcolor{green}{0.007} & -0.021 & -0.003 & -0.003 & 0.167 & 0.046 & 0.320 \\
            & 35-44	 & 0.008 & -0.013 &  \textcolor{green}{0.005}	& 0.003	& 0.003 & \textcolor{green}{-0.010} & \textcolor{red}{-0.007} & -0.003 &0.1600 & 0.047 & 0.313  \\
            & 45-54	 & 0.008 & -0.013 & \cellcolor{lightblue}{0.001} & \textcolor{red}{-0.005} & \textcolor{red}{0.001} & -0.013	& -0.003 & -0.006 & 0.136 & 0.020 & 0.326  \\
            & 55+ & 0.006 & \textcolor{red}{-0.018}	& -0.007 & \cellcolor{lightblue}{-0.002}	& 0.002	& -0.008 & -0.002 & \textcolor{red}{-0.010} & \textcolor{red}{0.100} & \textcolor{green}{0.060}	& 0.326 \\
        \bottomrule
    \end{tabular}
    }
    \label{tab:age_experiment_one}
\end{table*}

\begin{table*}
        \centering
    \caption{\textbf{Aggregate Shifts For Gender Attributes added in Isolated Context} across 4 models. The \textcolor{green}{green text} represents those values with significant ($p<0.05$) highest positive intensity and the \textcolor{red}{red text} represents those values with significant ($p<0.05$) highest negative intensity across each model's prediction per emotion. The \textcolor{blue}{blue highlighted cells} cells represent those attributes significantly similar ($p<0.05$) to the base state without any persona.}
            \resizebox{1.0\textwidth}{!}{
    \begin{tabular}{l l cc cc cc cc cc cc cc cc}
        \toprule
         \multirow{2}{*}{\textbf{Model}} & \multirow{2}{*}{\textbf{Attribute}} & \multicolumn{8}{c}{\textbf{Emotion}} & \multicolumn{3}{c}{\textbf{EPITOME}} \\ 
             & & anger & anticipation & disgust & fear & joy & sadness & surprise & trust & ER & IP & EX\\
             \midrule
             \multirow{4}{*}{Llama-3-70B} & male & \textcolor{red}{-0.005}	&\textcolor{red}{-0.001} &	-0.010 & \textcolor{red}{-0.014} & \textcolor{green}{0.002} & -0.009 & \textcolor{green}{0.009} &	\textcolor{green}{0.002} & 0.005&0.019&0.006\\
             &female & \cellcolor{lightblue}{\textcolor{green}{0.007}}	&\textcolor{red}{-0.001} & \cellcolor{lightblue}{-0.0005} & 0.004 &	-0.004 & \textcolor{green}{0.013} & 0.002 & 0.001 	& 0.108&0.052&0.029\\
             & non-binary & \cellcolor{lightblue}{-0.001} &	\textcolor{green}{0.015} &	\textcolor{red}{-0.017} & 0.018	&\textcolor{red}{-0.009} & -0.003 & \textcolor{red}{-0.007} &	\cellcolor{lightblue}{-0.001} & 0.061&0.036&0.046\\
            & gender-queer	 &\cellcolor{lightblue}{-0.001} & \cellcolor{lightblue}{0.001} &	\textcolor{green}{0.004}	& \textcolor{green}{0.0022} &	-0.003 & \textcolor{red}{-0.021}	&-0.003	& \textcolor{red}{-0.006} & 0.022&0.073&0.008\\
            \midrule
             \multirow{4}{*}{GPT-4o Mini} & male & \cellcolor{lightblue}{0.0004} &	0.007 &	\textcolor{red}{-0.010}	& -0.010 &-0.008 &\textcolor{green}{-0.013} &	\textcolor{green}{0.008} &	\textcolor{green}{0.004} &\textcolor{red}{-0.010} &	-0.006	&\textcolor{green}{-0.013} \\
             &female & \textcolor{red}{-0.012}	&\textcolor{red}{0.002} &	\textcolor{green}{-0.005} &	\textcolor{red}{-0.015} &	\textcolor{green}{-0.001}	&-0.026	&0.005 &	\textcolor{green}{0.004} &0.026	&\textcolor{red}{-0.013}&	-0.073 \\
             & non-binary & \cellcolor{lightblue}{0.001}	&0.005	&\textcolor{red}{-0.010}	&\textcolor{green}{-0.007} &	\textcolor{red}{-0.015} & -0.017 & 0.005 & \textcolor{red}{-0.003} & 0.043 & \textcolor{green}{0.006}	&-0.28 \\
            & gender-queer	 & \textcolor{green}{0.015}	&\textcolor{green}{0.008}	&-0.009	&\textcolor{green}{-0.007}	&-0.012 &	\textcolor{red}{-0.029}	&\textcolor{red}{0.000}	&-0.001&\textcolor{green}{0.050}	&-0.006&	\textcolor{red}{-0.300} \\
            \midrule
             \multirow{4}{*}{DeepSeek v3} & male & 0.021&	-0.006&	\cellcolor{lightblue}{0.000}&	\textcolor{red}{-0.022}&	\textcolor{red}{0.005}&	-0.027&	0.004&	\textcolor{red}{-0.015}&\textcolor{red}{-0.016}&	\textcolor{green}{0.040}&	-0.093\\
             &female & \textcolor{red}{0.011}&	\textcolor{red}{-0.011}&	\textcolor{green}{0.005}&	0.002&	-0.004&	-0.015&	\cellcolor{lightblue}{0.000}&	-0.011&0.033&	-0.006&	-0.166& \\
             & non-binary & 0.031&	-0.002&	\cellcolor{lightblue}{0.000}&	\cellcolor{lightblue}{0.000}&	-0.010&	-0.018&	-0.007&	-0.013&0.053&	-0.006&	-0.160 \\
            & gender-queer	 & \textcolor{green}{0.032}&	\textcolor{green}{0.003}&	\textcolor{red}{-0.002}&	\textcolor{green}{0.008}&	\textcolor{red}{-0.010}&	\textcolor{red}{-0.035}&	\textcolor{red}{-0.009}&	-0.014& \textcolor{green}{0.080}&	0.006&	-0.193\\
            \midrule
             \multirow{4}{*}{Gemini 2.0 Flash} & male & 0.019 &	-0.004 &	0.004	&0.002	& \cellcolor{lightblue}{\textcolor{green}{0.001}}	&-0.010&	0.001&	\cellcolor{lightblue}{0.001}& \textcolor{red}{0.070} &	\textcolor{red}{0.000} &	0.220 \\
             &female &0.0196 &	\textcolor{red}{-0.008} &	\textcolor{green}{0.007} &	\textcolor{green}{0.019}	&\cellcolor{lightblue}{0.000} & \textcolor{green}{-0.006} & \textcolor{red}{-0.004} &0.001 &0.086 &	\textcolor{green}{0.113}&	0.233\\
             & non-binary & \textcolor{green}{0.0206} &	0.010 &\cellcolor{lightblue}{-0.005} &	0.001	&-0.004	&\textcolor{red}{-0.026}	&\textcolor{green}{0.002}	&\textcolor{red}{-0.001} & \textcolor{green}{0.136}&	0.020	&0.240\\
            & gender-queer	 &\textcolor{red}{0.018}	&\textcolor{green}{0.0113}&	\textcolor{red}{-0.006}	&\textcolor{red}{-0.008} &	\cellcolor{lightblue}{-0.001}	&\textcolor{green}{-0.006}	& \cellcolor{lightblue}{0.000} & \textcolor{green}{0.003} &0.080 &	0.086	&0.173 \\
        \bottomrule
    \end{tabular}
    }
    \label{tab:gender_experiment_one}
\end{table*}

\section{Table of Shifts in Intersectional Attributes}\label{app:intersection-shift}
Tables \ref{tab:age_experiment_two}, \ref{tab:gender_experiment_two}
and \ref{tab:culture_experiment_two} represent those values of shift for the model where the attribute is added in the intersection with attributes from other demographic groups.  

\begin{table*}
        \centering
    \caption{\textbf{Aggregate Shifts For Cultural Attributes added in Intersections With Other Attributes} across 4 models. The \textcolor{green}{green text} represents those values with significant ($p<0.05$) highest positive intensity and the \textcolor{red}{red text} represents those values with significant ($p<0.05$) highest negative intensity across each model's prediction per emotion.}
    \resizebox{1.0\textwidth}{!}{
    \begin{tabular}{l l cc cc cc cc cc cc cc cc}
        \toprule
         \multirow{2}{*}{\textbf{Model}} & \multirow{2}{*}{\textbf{Attribute}} & \multicolumn{8}{c}{\textbf{Emotion}} & \multicolumn{3}{c}{\textbf{EPITOME}} \\ 
             & & anger & anticipation & disgust & fear & joy & sadness & surprise & trust & ER & IP & EX\\
             \midrule
             \multirow{9}{*}{Llama-3-70B} & Protestant Europe & -0.008 &	-0.018 &0.0144 &-0.011 & -0.002 & 0.010	&-0.008 &	0.001 & 0.001&0.0093&-0.122\\
             & English Speaking & -0.003	&\textcolor{green}{-0.006} & 	\textcolor{red}{0.003} &	-0.006 &	\textcolor{green}{0.001} &	\textcolor{red}{0.001} &	\textcolor{green}{0.000} &	0.000& 0.102&-0.013&0.181\\
             & Catholic Europe & -0.013 & \textcolor{red}{-0.028} & 0.018 & -0.008 &-0.003 &	0.020 &	-0.013 & \textcolor{green}{0.003} &0.071&0.056&-0.205\\
            & Confucian &\textcolor{red}{-0.041}	&-0.024 &	\textcolor{green}{0.022}	& \textcolor{red}{-0.019} &	\textcolor{red}{-0.011} &	\textcolor{green}{0.027} &	\textcolor{red}{-0.015} &	0.001 &-0.114&0.014&-0.512\\
            & West\&South Asia & -0.003 &	-0.010	& \textcolor{red}{0.003} &	-0.004 	&-0.002 &	0.010 &	-0.006 &	\textcolor{red}{-0.002} &0.039&0.048&-0.171\\
            & Latin America & \textcolor{green}{-0.001} &	-0.016 &	0.009	&-0.010	&-0.005	&0.003 &	-0.005	&0.003&0.023&	0.077&-0.173\\
            & African-Islamic & -0.007 &-0.019 	&0.010 & -0.008	&-0.006 & 0.003	&-0.010	&0.000 &0.054&0.016&-0.464\\
            & Orthodox Europe & -0.008 & -0.019 &	0.018	&\textcolor{green}{-0.002}	&-0.004 &	0.012 &	-0.012 & 0.002 &-0.004&0.0813&0.259\\
            \midrule
            \multirow{9}{*}{GPT-4o} & Protestant Europe & \textcolor{green}{0.004}	&-0.008 &	0.004	&-0.004&	\textcolor{green}{0.001}	&-0.004 &	-0.002 &	-0.001 &-0.002 &	0.003 &	-0.147 \\
             & English Speaking & 0.003 &	-0.006 & \textcolor{red}{0.002} &	-0.003 &	-0.001	&-0.003	&\textcolor{green}{-0.001}	& 0.000 &-0.008 &	0.004 &	\textcolor{green}{0.071} \\
             & Catholic Europe & 0.002 &	-0.009 & 0.008 &	-0.004	& \textcolor{green}{0.001}	&-0.001 &	-0.003 &	-0.001 &\textcolor{green}{-0.001}&	0.006&	-0.257\\
            & Confucian & \textcolor{red}{-0.012}&	\textcolor{red}{-0.010}	&\textcolor{green}{0.014}&	\textcolor{red}{-0.009} &	\textcolor{red}{-0.008}&	\textcolor{green}{0.012}	&\textcolor{red}{-0.005} &\textcolor{green}{0.001} &\textcolor{red}{-0.038} &	\textcolor{red}{-0.003}&	\textcolor{red}{-0.434} \\
            & West\&South Asia & 0.002	&-0.004 &	0.003 &	\textcolor{green}{-0.001} &	-0.002	&-0.001	&\textcolor{green}{0.000}&	-0.001&-0.021&	0.008 &	-0.249 \\
            & Latin America &0.002	&\textcolor{green}{-0.003} &	\textcolor{red}{0.002}	&\textcolor{green}{-0.001}&	-0.002	&\textcolor{green}{-0.001}	&-0.001 & -0.001& -0.015 &	\textcolor{green}{0.0139} &	-0.263\\
            & African-Islamic & 0.001 &	-0.009 &	0.004 &	-0.004 &	-0.002 &	\textcolor{red}{-0.007} &	-0.002  &-0.001 & -0.020	&-0.001 &	-0.374\\
            & Orthodox Europe & 0.001 &	-0.007	&0.006 &	-0.003 &	-0.002 &	0.001	&-0.003	&-0.001&-0.019&	0.004	&-0.253 \\
            \midrule
            \multirow{9}{*}{DeepSeek} & Protestant Europe & -0.005&	-0.009&	0.011&	\textcolor{red}{-0.007}&	0.001&	0.005&	-0.005&	-0.005& -0.204&	0.020&	-0.140&\\
             & English Speaking & -0.001&	\textcolor{red}{-0.001}&	0.004&	0.001&	0.002&	-0.004&	0.003&	-0.001& -0.021&	0.076&	\textcolor{green}{-0.021}&\\
             & Catholic Europe & \textcolor{red}{-0.021}&	-0.008&	0.002&	\textcolor{green}{0.0001}&	-0.003&	-0.001&	-0.004&	0.003& -0.210&	0.056&	-0.234&\\
            & Confucian & \textcolor{red}{-0.022}&	\textcolor{red}{-0.017}&	\textcolor{green}{0.025}&	0.001&	-0.008&	\textcolor{green}{0.039}&	-0.008&	\textcolor{red}{-0.014}& \textcolor{red}{-0.458}&	0.006&	-0.463&\\
            & West\&South Asia & 0.000&	-0.002&	0.003&	0.001&	0.001&	0.007&	-0.006&	-0.001& -0.136&	0.112&	-0.184&\\
            & Latin America & 0.002&	-0.001&	0.003&	0.002&	0.002&	0.003&	-0.001&	0.003& -0.207&	0.191&	-0.242&\\
            & African-Islamic & 0.000&	-0.010&	\textcolor{red}{0.000}&	-0.001&	-0.004&	0.010&	\textcolor{red}{-0.008}&	-0.006& -0.264&	\textcolor{green}{0.206}&	\textcolor{red}{-0.399}&\\
            & Orthodox Europe & -0.001&	-0.007&	0.011&	0.000&	0.000&0.009&	-0.005&	-0.002&-0.190&	0.066&	-0.227& \\
            \midrule
            \multirow{9}{*}{Gemini} & Protestant Europe & -0.007 &	\textcolor{green}{0.002} &	0.007 &	\textcolor{red}{-0.004}	&0.000 	&0.009	&-0.001&	\textcolor{green}{0.001} &-0.142 &	0.068 &	-0.224 \\
             & English Speaking & -0.002	&0.001 &	\textcolor{red}{-0.001} &	-0.002 &	0.000 &	\textcolor{red}{0.003} &	\textcolor{green}{0.004} &	0.000 &\textcolor{green}{0.055}	&0.025	&\textcolor{green}{-0.003}\\
             & Catholic Europe & -0.009	&-0.003 &	\textcolor{green}{0.010}	&\textcolor{green}{0.000} &	-0.002	&0.018	&\textcolor{red}{-0.005} &	-0.003 &-0.154	&0.187 &	-0.283\\
            & Confucian & \textcolor{red}{-0.034}	&\textcolor{red}{-0.006}	&0.001 & -0.001 &	\textcolor{red}{-0.006} &	\textcolor{green}{0.030} 	&-0.004 &	-0.001 &\textcolor{red}{-0.361}&	\textcolor{red}{0.004}	&\textcolor{red}{-0.421}\\
            & West\&South Asia & -0.008 &	-0.003 &	0.003 &	-0.003&	0.000 	&0.009	&-0.003 &	\textcolor{red}{-0.004} & -0.102 &	0.130	&-0.178 \\
            & Latin America &\textcolor{green}{0.000}	&-0.002 &	0.005 &	-0.002 &	-0.001	&0.006	&0.001	&-0.001 &-0.021 &	0.168	&-0.232 \\
            & African-Islamic & -0.007 &	-0.004 &	0.001 &	\textcolor{red}{-0.004}	&-0.003	&0.008	&\textcolor{red}{-0.005} &	0.000 &-0.199 &	\textcolor{green}{0.207} &	-0.348 \\
            & Orthodox Europe & -0.005 &	-0.002 &	\textcolor{green}{0.010} &	\textcolor{green}{0.000} &	\textcolor{green}{0.001} &	0.012	&-0.003	&-0.002 &-0.085&	0.063 &	-0.202 \\
        \bottomrule
    \end{tabular}
    }
    \label{tab:culture_experiment_two}
\end{table*}

\begin{table*}
        \centering
    \caption{\textbf{Aggregate Shifts For Age Attributes added in Intersections With Other Attributes} across 4 models. The \textcolor{green}{green text} represents those values with significant (p<0.05) highest positive intensity and \textcolor{red}{red text} represents those values with the significant (p<0.05)  highest negative intensity across each model's prediction per emotion.}
        \resizebox{1.0\textwidth}{!}{
    \begin{tabular}{l l cc cc cc cc cc cc cc cc}
        \toprule
         \multirow{2}{*}{\textbf{Model}} & \multirow{2}{*}{\textbf{Attribute}} & \multicolumn{8}{c}{\textbf{Emotion}} & \multicolumn{3}{c}{\textbf{EPITOME}} \\ 
             & & anger & anticipation & disgust & fear & joy & sadness & surprise & trust & ER & IP & EX\\
             \midrule
             \multirow{6}{*}{Llama-3-70B} & 0-17 & \textcolor{green}{0.015}	&\textcolor{green}{0.012}&	\textcolor{green}{-0.005} &	\textcolor{green}{0.013}&	\textcolor{green}{0.003} &\textcolor{green}{0.005}	&\textcolor{green}{0.015}&	-0.001&0.160&0.055&0.096\\
             & 18-24 & 0.001	&0.004	&0.000	&0.000	& 0.000	&\textcolor{green}{-0.005} &	0.003 &	\textcolor{red}{-0.002} &0.051&0.015&0.071&\\
             & 25-34 & -0.001	&-0.001 &	0.001&	-0.004 &	0.001 &	-0.002 &	-0.001 & 0.000 &0.02&0.002&0.069\\
            & 35-44	 & -0.004 &	-0.006 &	0.003 &	-0.010 &	-0.001	&-0.002 &	-0.005 & 0.002 &0.017&-0.0011&0.022\\
            & 45-54	 &-0.009 &	-0.009 &	0.004 &	-0.010 &	-0.001	&-0.001 &	-0.006 & \textcolor{green}{0.003} &0.035&0.0035&-0.005&\\
            & 55+ &\textcolor{red}{-0.019} &	\textcolor{red}{-0.021} &	\textcolor{red}{0.007}	&\textcolor{red}{-0.018}	&\textcolor{red}{-0.003}	&0.003	&\textcolor{red}{-0.010}	&-0.001 & 0.121 &0.077&0.015& \\
            \midrule
             \multirow{6}{*}{GPT-4o} & 0-17 & -0.006	&\textcolor{green}{0.002} & \textcolor{red}{-0.004} &	\textcolor{green}{0.009}	&0.003 & -0.002 &\textcolor{green}{0.007}	&\textcolor{green}{0.000}&\textcolor{green}{0.046}&	-0.002&	\textcolor{green}{0.023}\\
             & 18-24 & \textcolor{green}{0.003} &	-0.002	&-0.002 &	-0.002	&\textcolor{green}{0.005}	&\textcolor{red}{-0.009} &	0.006 &	\textcolor{red}{-0.003} &0.018 &	0.005 &	0.118 \\
             & 25-34 & 0.000	&-0.001	&0.002 & -0.002 &	0.002 &	-0.002&	0.004	&-0.002& 0.019	&0.005	&0.119 \\
            & 35-44	 & -0.001 &	-0.007 & \textcolor{green}{0.003} &	-0.005	&\textcolor{red}{0.000} &-0.001	& 0.000 &	-0.002 & 0.018 &	0.005 &	0.119\\
            & 45-54	 & -0.004 &	-0.011  & 0.002 &	-0.006 &	0.001 &	0.003 &	-0.003	&-0.002 &0.018&	0.005 &	0.119\\
            & 55+ & \textcolor{red}{-0.007} &	\textcolor{red}{-0.018} &	0.000 &	\textcolor{green}{-0.007} &	0.002 &	\textcolor{green}{0.007}  &\textcolor{red}{-0.005}	&\textcolor{red}{-0.003} & 0.025 &	-0.002	&\textcolor{red}{-0.027} \\
            \midrule
             \multirow{6}{*}{DeepSeek} & 0-17 & 0.001&	\textcolor{green}{0.019}&	\textcolor{red}{-0.008}&	\textcolor{green}{0.022}&	\textcolor{red}{-0.003}&	-0.009&	\textcolor{green}{0.015}&	\textcolor{green}{0.006}& \textcolor{green}{0.203}&	0.039&	0.055&
\\
             & 18-24 & \textcolor{green}{0.003}&	0.011&	-0.003&	0.003&	0.003&	\textcolor{red}{-0.015}&	0.010&	0.004& 0.106&	0.064&	0.102&\\
             & 25-34 & 0.002&	0.004&	0.001&	-0.001&	0.004&	-0.008&	0.004&	0.002& 0.058&	0.023&	\textcolor{green}{0.124}&\\
            & 35-44	 & 0.001&	-0.002&	\textcolor{green}{0.003}&	-0.003&	0.005&	-0.003&	0.003&	0.001& -0.005&	0.024&	0.075&\\
            & 45-54	 & 0.000&	\textcolor{red}{-0.005}&	\textcolor{green}{0.003}&	\textcolor{red}{-0.006}&	\textcolor{green}{0.006}&	\textcolor{red}{-0.001}&	0.001&	0.000& -0.005&	0.013&	0.082&\\
            
            & 55+ & \textcolor{red}{-0.007}&	-0.012&	0.001&	-0.004&	0.004&	\textcolor{green}{0.007}&	\textcolor{red}{-0.001}&	\textcolor{red}{-0.003}& -0.001&	0.0188&	0.015&\\
            \midrule
             \multirow{6}{*}{Gemini} & 0-17 & \textcolor{red}{-0.007}  &	\textcolor{green}{0.017} &	\textcolor{red}{-0.017} &	\textcolor{green}{0.025} &	0.001 &	-0.001 &	\textcolor{green}{0.017}	&\textcolor{green}{-0.001} &\textcolor{green}{0.163} &	\textcolor{red}{-0.069} &	0.077\\
             & 18-24 & 0.0000	& 0.009 &	-0.005	&0.003	&0.001 &	\textcolor{green}{-0.003}	&0.007 &	-0.003 &0.125&	-0.058	&0.072\\
             & 25-34 & \textcolor{green}{0.008} & 0.000 & \textcolor{green}{0.003} &	0.003&	\textcolor{green}{0.002}&	-0.001 &	0.001	&-0.002 &0.118	&-0.066 &	\textcolor{green}{0.114} \\
            & 35-44	 & 0.005	&0.000 &	0.000 &	-0.001 &	\textcolor{red}{-0.001} &	-0.001	&0.001 &	-0.002 & 0.111 &	-0.063	&0.097\\
            & 45-54	 & 0.001	&\textcolor{red}{-0.005}	&0.002	&\textcolor{red}{-0.002}	&0.000&	0.004 &	0.000 & -0.003 &0.083 &	-0.053 &	0.078\\
            & 55+ & -0.003 &	-0.011 &	0.002	&\textcolor{red}{-0.002}	&\textcolor{red}{-0.001} &	\textcolor{green}{0.011}	&\textcolor{red}{-0.001} &	\textcolor{red}{-0.005} & \textcolor{red}{0.080}	&\textcolor{green}{-0.019} &	\textcolor{red}{0.031}\\
        \bottomrule
    \end{tabular}
    }
    \label{tab:age_experiment_two}
\end{table*}

\begin{table*}
        \centering
    \caption{\textbf{Aggregate Shifts For Cultural Attributes added in Intersections With Other Attributes} across 4 models. The \textcolor{green}{green text} represents those values with significant ($p<0.05$) highest positive intensity and \textcolor{red}{red text} represents those values with the significant ($p<0.05$) highest negative intensity across each model's prediction per emotion.}
            \resizebox{1.0\textwidth}{!}{
    \begin{tabular}{l l cc cc cc cc cc cc cc cc}
        \toprule
         \multirow{2}{*}{\textbf{Model}} & \multirow{2}{*}{\textbf{Attribute}} & \multicolumn{8}{c}{\textbf{Emotion}} & \multicolumn{3}{c}{\textbf{EPITOME}} \\ 
             & & anger & anticipation & disgust & fear & joy & sadness & surprise & trust & ER & IP & EX\\
             \midrule
             \multirow{4}{*}{Llama-3-70B} & male & \textcolor{green}{0.003} &	\textcolor{red}{0.001} &	-0.005	&\textcolor{red}{-0.014}&	0.001 &	 \textcolor{red}{-0.013}	&\textcolor{red}{-0.002}&	\textcolor{green}{0.001} &\textcolor{red}{0.038} 	&-0.017 &\textcolor{red}{-0.004} \\
             &female & \textcolor{red}{-0.006}&	0.002 &	\textcolor{green}{-0.001}	&\textcolor{green}{0.010} &	\textcolor{red}{-0.001}	&0.005 &	0.000&	\textcolor{green}{0.001} &\textcolor{green}{0.148}&	\textcolor{red}{-0.006}&	0.091 \\
             & non-binary &-0.005 &	0.011&	\textcolor{red}{-0.006} &	0.003	&0.001 &	-0.004	& \textcolor{green}{0.001}	&-0.002&0.101	& \textcolor{green}{0.074}	&0.091 \\
            & gender-queer	 &-0.000	&\textcolor{green}{0.013} &	-0.005&	0.005 &	\textcolor{green}{0.003} &	-0.010 &	\textcolor{green}{0.001}&	\textcolor{red}{-0.005} &0.101 &	\textcolor{green}{0.074} &	0.091\\
            \midrule
             \multirow{4}{*}{GPT-4o} & male & \textcolor{green}{0.004}	 &-0.001	 &\textcolor{red}{-0.001} &\textcolor{red}{-0.003}	 &-0.002	 &-0.005 &	\textcolor{green}{0.001} &	-0.001 & \textcolor{red}{0.007} &	-0.001 &	\textcolor{green}{0.045}\\
             &female & 0.005 &	\textcolor{red}{-0.003}	 &\textcolor{green}{0.002} &	0.004 &	\textcolor{green}{0.000}	 &\textcolor{green}{-0.001}	 &\textcolor{red}{-0.004} &	-0.001 & \textcolor{green}{0.035}&	-0.001&	0.040\\
             & non-binary & 0.008	 &0.005	 &0.001 &	0.003 &	\textcolor{red}{-0.004}	 &-0.005	 &-0.001 &	-0.001 & 0.021&	\textcolor{red}{-0.008	}&\textcolor{green}{-0.032}\\
            & gender-queer	 & \textcolor{green}{0.014} &	\textcolor{green}{0.008} &	0.000 &	\textcolor{green}{0.005}	 &\textcolor{red}{-0.004}	 &\textcolor{red}{-0.009}	 &-0.002	 &-0.001 & 0.027	&-0.005	&-0.018\\
            \midrule
             \multirow{4}{*}{Deep Seek} & male & 0.005&	\textcolor{red}{0.000}&	\textcolor{red}{-0.005}&	\textcolor{red}{-0.009}&	\textcolor{green}{0.000}&	\textcolor{red}{-0.013}&	\textcolor{red}{-0.003}&	-0.002& 0.011&	0.010&	0.018& \\
             &female & \textcolor{red}{0.000} &	\textcolor{red}{0.002}&	\textcolor{green}{0.000}&	\textcolor{green}{0.013}&	-0.004&	0.000&	0.000&	0.001& 0.067&	0.095&	0.006\\
             & non-binary & 0.004&	0.002&	-0.002&	0.003&	-0.004&	-0.003&\textcolor{green}{0.002}&	\textcolor{green}{0.002}& 0.051&	0.0255&	0.071\\
            & gender-queer	 & \textcolor{green}{0.009}&	\textcolor{green}{0.005}&	-0.002&	0.007&	\textcolor{red}{-0.005}&	-0.003&	0.001&	0.000& 0.027&	0.045&	0.045\\
            \midrule
             \multirow{4}{*}{Gemini} & male & 0.005 &	\textcolor{red}{0.001} &	0.003 &	-0.002	&\textcolor{green}{-0.001 }&	-0.003	&\textcolor{red}{0.002}	&\textcolor{red}{0.000} &\textcolor{red}{-0.031}	&\textcolor{red}{0.022}	&-0.025\\
             &female & \textcolor{red}{0.004} &	0.004 &	\textcolor{green}{0.004} &	\textcolor{green}{0.008} &	-0.002	&\textcolor{green}{0.003}	&\textcolor{red}{0.002}	&-0.001 & \textcolor{green}{0.056}	&0.055 &	\textcolor{green}{0.004}\\
             & non-binary & 0.006	&0.006	&-0.002	&-0.002 &	-0.002 &	-0.003 &	0.003	&0.001 & -0.011 &	0.055	&\textcolor{red}{-0.034}\\
            & gender-queer	 & \textcolor{green}{0.007} &	\textcolor{green}{0.011} &	\textcolor{red}{-0.003} &	-0.002	&-0.002	&\textcolor{red}{-0.006}	&\textcolor{green}{0.004}	&\textcolor{green}{0.002} &\textcolor{red}{-0.031}&	\textcolor{green}{0.080}	&-0.016 \\
        \bottomrule
    \end{tabular}
    }
    \label{tab:gender_experiment_two}
\end{table*}

\section{Results: Personas recalled in Base state}\label{app:persona-base}

To interpret the model's \texttt{base state}, we characterize the categories of the persona output in the table \ref{tab:persona-base}. We see that the personas recalled by the base are limited to Profession, Behavioural and Topic Related categories.

\begin{table}
\centering
\caption{\textbf{Examples of Personas} Recalled in the Base State}
\label{tab:persona-base}
\resizebox{0.48\textwidth}{!}{
\begin{tabular}{@{}ll@{}}
\toprule
\textbf{Category} & \textbf{Example} \\
\midrule
\multirow{4}{*}{Profession-Related} 
    & A nursing student accused of reporting someone for cheating \\
    & A student \\
    & A teacher or educator \\
    & An experienced scuba diver or thrill-seeker \\
\midrule
\multirow{4}{*}{Behavioural}  
    & A concerned individual \\
    & A skeptical individual \\
    & A person from a stable background who values nostalgia \\
    & A concerned and empathetic individual \\
\midrule
\multirow{4}{*}{Topic Related}  
    & A person who is frustrated with their neighbor's behavior \\
    & A person who has recently completed their B.Sc degree \\
    & A student seeking a recommendation letter \\
    & A victim of rumours and gossip \\
\bottomrule
\end{tabular}
}
\end{table}

. 


\section{Results: Topic to Attribute Ratio}\label{app:topic-attribute-ratio}

We devise the Topic to Attribute Variance (TAV) Ratio as the following:

\begin{equation}
\resizebox{0.45\textwidth}{!}{$
    TAV = \frac{\text{Variance of the attribute's embeddings from the base culture}}%
    {\text{Variance of the attribute's embeddings from the attribute's mean}}
$}
\end{equation}
A ratio > 1 indicates that the response embeddings for the given attribute are centered around the attribute's characteristics and stereotypes, while a ratio < 1 reflects that the response is more likely to be topic dependent.  We show the topic to attribute variance in Table \ref{tab:model-cultural-alignment}.

\begin{table}[]
        \centering
    \caption{\textbf{Topic to Attribute Ratio} is calculated to assess whether the model's response for the attribute is skewed towards the topic or the attribute's characteristic. The values highlighted in \textcolor{blue}{blue} represent those attributes where the model is likely to generate a response focusing on its characteristic.}
        \resizebox{0.45\textwidth}{!}{
    \begin{tabular}{l cccc }
        \toprule
        \multirow{2}{*}{\textbf{Persona}} &\multicolumn{4}{c}{\textbf{Topic to Attribute Ration}} \\
        & GPT-4o-Mini & Llama-70B & Gemini 2.0 & DeepSeek \\
        \midrule
        0-17 & 0.708 & 0.778 & 0.816& 0.886 \\
        18-24 & 0.700 & 0.714 & 0.778 & 0.839 \\
        25-34 & 0.636 &0.688 & 0.749 & 0.806 \\
        35-44 & 0.640 &0.685 & 0.751 & 0.803 \\
         45-54 & 0.657  &0.694 & 0.747 & 0.812\\
          55+ & 0.693 &0.726 & 0.762 & 0.839\\
       male& 0.603 & 0.690 & 0.751 & 0.804 \\
       female& 0.656 & 0.721 & 0.750 & 0.820\\
       non-binary& 0.717 & 0.724 & 0.754 & 0.867\\
       gender-queer& 0.786 & 0.819 & 0.823 &\cellcolor{lightblue}{1.012}\\
       Protestant Europe &0.628 &0.834 & 0.838 & 0.942\\
       English Speaking & 0.566 &0.729 & 0.803 & 0.820\\
       Catholic Europe& 0.691 &0.891 & 0.910& 0.999\\
        Confucian& \cellcolor{lightblue}{1.021} &\cellcolor{lightblue}{1.589} &\cellcolor{lightblue}{1.137} & \cellcolor{lightblue}{1.227}\\
        West and South Asia& 0.682 &0.797 & 0.866& 0.934\\
      Latin America& 0.715 & 0.898 & 0.946 & \cellcolor{lightblue}{1.021}\\
      African-Islamic& 0.779 & \cellcolor{lightblue}{1.01} & 0.961 &\cellcolor{lightblue}{1.093} \\
      Orthodox Europe& 0.678 &0.845 & 0.873& 0.945\\
        \bottomrule
    \end{tabular}
    }
    \label{tab:model-cultural-alignment}
\end{table}




\section{Real World Gallup Scores}

We collect the emotion scores from the Emotion World Report ~\cite{gallupEmotion}, which presents a comprehensive study across 142 nations. To process this dataset into our cultures, we prepare a mapping according to the Inglehart–Welzel Cultural Map from the countries to culture. We then decompose the emotion words studied in the Gallup poll into Plutchik's 8 Basic emotions \cite{plutchik1980general} which are used by NRC and perform an aggregated weighted emotion score for each culture as follows in Table \ref{tab:gallup_emotion_scores}. 

As seen in Table \ref{tab:gallup_emotion_scores}, the intensities for disgust and surprise are $0$ across all cultures, and this is  due to the absence of these intensities in the original dataset ~\cite{gallupEmotion}

\begin{table}
\centering
\caption{\textbf{Real World Affective Scores} for emotion categories across cultures according to the Gallup World Poll}
\resizebox{0.5\textwidth}{!}{
\begin{tabular}{lcccccccc}
\toprule
\textbf{Culture} & \textbf{Anger} & \textbf{Anticipation} & \textbf{Disgust} & \textbf{Fear} & \textbf{Joy} & \textbf{Sadness} & \textbf{Surprise} & \textbf{Trust} \\
\midrule
Protestant Europe     &     0.008	&0.103&	0.0&	0.048&	0.182&	0.056&	0.0	&0.113   \\
English Speaking       &     0.013	&0.104&	0.0&	0.060&	0.180&	0.069&	0.0	&0.112      \\
Catholic Europe     &     0.011&	0.100	&0.0&	0.054&	0.170&	0.062	&0.0	&0.107      \\
Confucian     &      0.011&	0.090&	0.0&	0.043&	0.166&	0.043&	0.0&	0.097   \\
West and South Asia      &      0.014&	0.102&	0.0	&0.046	&0.182&	0.060&	0.0	&0.112    \\
Latin America    &      0.013	&0.111&	0.0&	0.068&	0.190&	0.082&	0.0	&0.118    \\
African-Islamic       &    0.023&	0.093&	0.0&	0.068&	0.154&	0.082&	0.0	&0.100      \\
Orthodox Europe       &      0.016	&0.094&	0.0	&0.054&	0.151	&0.068&	0.0&	0.104     \\
\bottomrule
\end{tabular}
}
\label{tab:gallup_emotion_scores}
\end{table}

\end{document}